\documentclass[runningheads]{llncs}

\usepackage{eccv}

\usepackage{eccvabbrv}

\usepackage{graphicx}
\usepackage{booktabs}

\usepackage[pagebackref,breaklinks,colorlinks,citecolor=eccvblue]{hyperref}

\usepackage[utf8]{inputenc} 
\usepackage[T1]{fontenc}    
\usepackage{booktabs}       
\usepackage{amsfonts}       
\usepackage{nicefrac}       
\usepackage{microtype}      
\usepackage{color}
\usepackage{bigstrut}
\usepackage{hhline}
\usepackage{colortbl}
\usepackage{tabularx}
\usepackage{subcaption}
\usepackage{arydshln}
\usepackage{array,multirow}
\usepackage{xfrac}
\usepackage{bm}
\usepackage{soul}
\usepackage[export]{adjustbox}
\usepackage{graphicx}
\usepackage{verbatim}
\usepackage{mwe}
\usepackage[shortcuts]{extdash}
\usepackage[toc,page]{appendix}
\usepackage{enumitem}
\usepackage{bbold}
\usepackage[normalem]{ulem}
\usepackage[dvipsnames]{xcolor}
\usepackage{bbm}
\usepackage{ulem}
\usepackage{float}
\usepackage{enumitem} 
\usepackage{algorithm}
\usepackage{algorithmic}
\usepackage{caption}
\usepackage[font={small}]{caption}
\usepackage{pdfpages}

\usepackage[detect-all]{siunitx}
\usepackage{etoolbox}

\usepackage{mathtools, nccmath}

\newcolumntype{L}[1]{>{\raggedright\let\newline\\\arraybackslash\hspace{0pt}}m{#1}}
\newcolumntype{R}[1]{>{\raggedleft\let\newline\\\arraybackslash\hspace{0pt}}m{#1}}
\newcolumntype{C}[1]{>{\centering\let\newline\\\arraybackslash\hspace{0pt}}m{#1}}

\begin{document}

\title{Toward INT4 Fixed-Point Training via Exploring Quantization Error for Gradients}
\titlerunning{INT4 Fixed-Point Training}

\author{Dohyung Kim\inst{1} \and
Junghyup Lee\inst{1} \and
Jeimin Jeon \inst{1,2} \and \\
Jaehyeon Moon \inst{1,2} \and
Bumsub Ham \inst{1,}\thanks{Corresponding author}
}

\authorrunning{D.~Kim et al.}

\institute{$^1$Yonsei University, $^2$Articron Inc.\\
\url{http://cvlab.yonsei.ac.kr/projects/LBT}}
\maketitle
\vspace{-6mm}

\begingroup
\renewcommand{\thefootnote}{}
\endgroup

\setcounter{footnote}{0}
\renewcommand{\thefootnote}{\arabic{footnote}}

\begin{abstract}
Network quantization generally converts full-precision weights and/or activations into low-bit fixed-point values in order to accelerate an inference process. Recent approaches to network quantization further discretize the gradients into low-bit fixed-point values, enabling an efficient training. They typically set a quantization interval using a min-max range of the gradients or adjust the interval such that the quantization error for entire gradients is minimized. In this paper, we analyze the quantization error of gradients for the low-bit fixed-point training, and show that lowering the error for large-magnitude gradients boosts the quantization performance significantly. Based on this, we derive an upper bound of quantization error for the large gradients in terms of the quantization interval, and obtain an optimal condition for the interval minimizing the quantization error for large gradients. We also introduce an interval update algorithm that adjusts the quantization interval adaptively to maintain a small quantization error for large gradients. Experimental results demonstrate the effectiveness of our quantization method for various combinations of network architectures and bit-widths on various tasks, including image classification, object detection, and super-resolution.
\vspace{-0.2cm}
\end{abstract}

\vspace{-0.8cm}
\section{Introduction}\label{sec:intro}
\vspace{-0.3cm}
Over the past decade, convolutional neural networks (CNNs) have shown the effectiveness on various applications in computer vision~\cite{deng2009imagenet,lin2014microsoft,everingham2010pascal}. The networks exploit wide~\cite{szegedy2015going} and deep architectures~\cite{he2016deep,xie2017aggregated} with lots of training samples for better performance, which requires a large amount of memory to store,~\eg,~weights, activations, and/or gradients, typically using full-precision values. Multiply-accumulate operations (MACs) with full-precision values are computationally demanding for both training and inference processes. Network quantization alleviates this problem by replacing the full-precision values with low-bit fixed-point ones (\ie, integer format). This allows to employ an efficient integer arithmetic, while reducing the required memory and computational cost simultaneously. Recent studies focus on quantizing weights and/or activations in a forward pass to accelerate an inference process. Several methods have shown that the bit-width could be reduced to extremely low ones, \eg, 3-bit, while retaining the accuracy of an original model~\cite{jung2019learning,gong2019differentiable,esser2019learned,lee2021ewgs,kim2021daq,yang2019quantization}. They, however, also require high computational cost at training time, since gradients for backward propagation are kept to full-precision values. For an efficient training process, quantizing the gradients into low-bit widths is crucial, while minimizing the performance drop.

Low-bit training approaches reduce the bit-width of gradients for an efficient backward propagation, which can be categorized into low-bit floating-point (FLP) and fixed-point (FXP) training methods. Low-bit FLP methods, representing the gradients with low-bit FLP values, have been widely used to boost the efficiency for backward propagation, but they still adopt MACs with FLP values~\cite{wang2018training,sun2019hybrid,cambier2020shifted,koster2017flexpoint}. Low-bit FXP methods have recently attracted significant attention that enable using integer arithmetic operations for backward propagation~\cite{zhou2016dorefa,zhu2020towards,liu2022IQB}. To this end, they exploit a discrete quantization function that maps full-precision gradients~(\eg,~with 32-bit FLP values) into low-bit FXP ones. Specifically, the quantization function first normalizes the full-precision gradients within a quantization interval, and then maps them to the low-bit ones using a discretizer (\ie, rounding function). Since finding an optimal quantization interval brings better quantization performance, recent approaches to quantizing weight and/or activation propose to learn the interval end-to-end, providing state-of-the-art results~\cite{gong2019differentiable,esser2019learned,lee2021ewgs,choi2018pact}. Adopting the learnable interval to quantize gradients is however computationally intractable, mainly due to computing derivatives of gradients (\ie, Hessian). For this reason, current FXP methods simply set the interval to the min-max range of gradients~\cite{zhou2016dorefa,fournarakis2021hindsight}. A recent work proposes to adjust the interval such that the quantization error for entire gradients is minimized~\cite{zhu2020towards}. We have found that this approach narrows the quantization interval significantly, compared to the methods using a min-max range, since most gradients are distributed around a zero value~\cite{zhu2020towards}, while the min-max range spanning entire gradients is relatively very wide. Narrowing the quantization interval drastically leads to a significant quantization error for large gradients around a tail of distribution that have larger magnitudes affecting the training process dominantly~\cite{lee2021ewgs,ke2017lightgbm}. 



In this paper, we introduce a simple yet effective method for a low-bit FXP training that updates the quantization interval for gradient quantization in a way of maintaining a small quantization error for large gradients. We conjecture that minimizing the quantization error for entire gradients causes a significant error for large gradients, which leads to an unstable training process. Our approach instead lowers the quantization error for large gradients in the FXP training. To this end, we derive an upper bound of the quantization error for the large gradients using a quantization interval, and obtain a condition for the interval that lowers the upper bound of the quantization error. Based on this condition, we propose an interval update algorithm that adjusts the quantization interval adaptively, maintaining a low quantization error for large gradients accordingly. We apply our method to various network architectures with different bit-widths, and achieve superior results on various vision tasks including image classification, object detection, and super-resolution. The main contributions of our work can be summarized as follows:
\begin{itemize}
  \item[$\bullet$] We have found that minimizing the quantization error for entire gradients enlarges the quantization error for large gradients. Based on this, we propose to focus on reducing the quantization error for large gradients that play an important role for the low-bit FXP training.
  \item[$\bullet$] We derive an upper bound of the quantization error for large gradients, and compute an optimal condition for quantization intervals lowering the quantization error for large gradients. We design an interval update algorithm for the low-bit FXP training with a negligible computational overhead.
  \item[$\bullet$] We demonstrate the effectiveness of our approach to updating the interval to maintain a small quantization error for large gradients with various architectures on standard benchmarks especially in 4-bit setting, and show an extensive analysis of our method.
\end{itemize}
\vspace{-6mm}
\section{Related work}
\vspace{-0.3cm}
\subsection{Low-bit FLP training}
\vspace{-0.2cm}

FLP training approaches accelerate a training process by lowering the bit-width of gradients into a 16-bit~\cite{koster2017flexpoint,micikevicius2017mixed} or an 8-bit~\cite{wang2018training,sun2019hybrid,cambier2020shifted}. A FLP value consists of exponent and mantissa parts, which represent dynamic range and precision, respectively. The FLP training approaches carefully assign bit-widths for the exponent and mantissa parts in order to minimize an accuracy drop caused by gradient quantization. Specifically, the work of~\cite{wang2018training} shows in-depth studies on distributions of weights, activations, and gradients, and proposes to use an 8-bit FLP value. More specifically, it uses 1, 5, and 2-bits for sign, exponent, and mantissa parts, respectively, in forward and backward propagations. After that, several approaches apply different formats of the 8-bit FLP value for weights, activations, and gradients~\cite{sun2019hybrid}, or leverage scaling and shifting operations to adjust gradients within the dynamic range for an 8-bit FLP value~\cite{cambier2020shifted}. More recently, the work of~\cite{sun2020ultra} adopts a radix-4 data format specialized for the FLP training with 4-bit values. Current FLP training approaches have shown the effectiveness to the low-bit training, but they still require MACs with FLP values at training time. On the contrary, our work is for the FXP training that is more hardware-friendly in terms of computational power and chip area, compared to the FLP training~\cite{horowitz20141,zhang2020fixed}.

\vspace{-0.3cm}
\subsection{Low-bit FXP training}
\vspace{-0.2cm}
Using FXP gradients for backward propagation degrades the performance significantly compared to the FLP counterparts within the same bit-width, due to the narrow dynamic range compared to that of FLP values~\cite{zhong2022exploring}. MACs with FLP values are more resource-intensive than those of FXP values, suggesting that a FXP training is suitable for hardware implementation~\cite{horowitz20141,zhang2020fixed}. In this regard, recent works have focused on the FXP training that converts full-precision gradients into low-bit FXP values~\cite{zhou2016dorefa,fournarakis2021hindsight,zhu2020towards,zhang2020fixed,zhao2021distribution}. DorefaNet~\cite{zhou2016dorefa} quantizes gradients into FXP values for the first time, and adopts the stochastic rounding technique~\cite{gupta2015deep} to reduce the average quantization error in the training process. FPT~\cite{zhang2020fixed} proposes to assign different bit-widths for each layer, and changes them continually during training. FQT~\cite{chen2020statistical} presents a per-sample quantization approach by employing multiple quantizers for different samples within a batch. Although this effectively captures the dynamic range variations across samples, extra FLP operations are required to normalize each sample, which is less efficient compared to layer-wise quantization techniques. More recently, IQB~\cite{liu2022IQB} introduces a piecewise FXP format for gradient quantization that lowers the quantization error effectively, while avoiding clipping gradients. However, the piecewise format requires specially designed hardwares. On the contrary, our method uses a layer-wise quantization with a uniform quantizer, which aligns well with hardware implementation, while boosting the quantization performance significantly in low-bit FXP training.

Closely related to ours, several methods~~\cite{zhu2020towards,zhao2021distribution} design quantizers for gradients considering the quantization error. DSGC~\cite{zhu2020towards} claims that minimizing the quantization error for entire gradients is important for low-bit FXP training. It thus proposes to search the quantization interval that maximizes cosine similarity between full-precision and quantized gradients. This approach makes the quantization interval significantly narrow, since the majority of gradients are concentrated near zero. The narrow interval incurs substantial quantization error for large gradients that affect the training process dominantly. In contrast to DSGC, we design an update algorithm adjusting the quantization interval adaptively to lower the quantization error for large gradients. DAIQ~\cite{zhao2021distribution} employs a channel-wise quantization strategy, using multiple quantizers with different quantization intervals along the channel dimensions of gradients for each layer, in order to reduce the quantization error effectively. However, this approach is less suitable for hardware implementation compared to a layer-wise quantization method. DAIQ also designs a magnitude-aware clipping strategy that lowers the quantization error weighted by gradient magnitudes. It sets a clipping value as the running mean of the maximum gradient over training iterations. DAIQ applies this technique to the channels whose gradients follow inverted-T distributions. Otherwise, it employs the min-max quantizer. Different from DAIQ, our approach adopts a layer-wise quantization method exploiting a single quantizer per layer, which is more feasible for hardware implementation, and efficient in terms of computational cost. Moreover, our quantizer is applicable to the gradients of any distributions, enabling lowering the quantization error for large gradients regardless of their distributions.
\vspace{-0.3cm}
\section{Method}
\vspace{-0.2cm}
\subsection{Overview}
\vspace{-0.1cm}
\label{sec:preliminary}
Following recent works~\cite{zhou2016dorefa,fournarakis2021hindsight,zhu2020towards,zhang2020fixed}, we quantize full-precision weights, activations, and gradients into low-bit FXP ones. To this end, we use a uniform quantizer that converts a full-precision input~$x$ (\ie, weights, activations, or gradients) to a $b$-bit quantized output. We adopt a layer-wise quantization for efficiency. Specifically, we clip the input within a quantization interval, parameterized by a clipping value~$c$, and normalize it using a scale factor~$s$ to obtain a normalized input~$x_n$ as follows:
  \begin{equation}
  \small
      \label{eq:normalization} 
      x_n = \frac{\mathrm{clip}(x,c)}{s},
  \end{equation}
where the clipping function and the scale factor are defined differently depending on distributions of inputs~\cite{lee2021ewgs,kim2021daq,esser2019learned,zhu2020towards}. For example, for the input data with a zero-centered distribution,~\eg,~weights or gradients, the clipping function and the scale factor are designed as 
  \begin{equation}
  \small
  	\mathrm{clip}(x,c)=\min(\max(x,-c),c),~~~~~s = \frac{c}{2^{b-1}-1}.
  \end{equation}
Differently, they are defined as 
  \begin{equation}
  \small
  	\mathrm{clip}(x,c)=\min(\max(x,0),c),~~~~~s = \frac{c}{2^{b}-1},
  \end{equation}
if the input data follows a half-normal distribution, \eg,~activations after a ReLU. We then obtain a quantized output~$Q(x)$ by applying a rounding~$\lceil\cdot\rfloor$ to the normalized input~$x_n$, followed by multiplying it with the scale factor~$s$ for denormalization as follows:
  \begin{equation}
  \small
      \label{eq:quantizer}
      Q(x) = s\left\lceil x_n \right\rfloor.
  \end{equation}
Following the works of~\cite{gong2019differentiable,esser2019learned,lee2021ewgs,choi2018pact}, we learn the clipping values~$c$ (\ie,~the quantization interval) end-to-end for weight and activation quantizers at each layer. Note that learning the clipping values for a gradient quantizer is intractable, since it requires to compute the derivatives of gradients~(\ie,~Hessian). We manually set the clipping value for gradients, denoted by~$c_g$, to~$\gamma g_{max}$, where $\gamma \in (0,1]$ is a clipping factor, and $g_{max}$ is the maximum absolute gradient~(\ie,~$max(\vert G\vert)$, where we denote by~$G$ a set of entire gradients in a single layer). Note that the clipping factor~$\gamma$ controls the quantization interval. For example, the interval becomes narrow as the clipping factor decreases. Previous methods set the clipping factor to 1, suggesting that all gradients are taken into account to estimate the quantization interval~\cite{zhou2016dorefa}, or adjust the factor to minimize the quantization error for entire gradients~\cite{zhu2020towards}. In contrast to these approaches, we propose to update the clipping factor adaptively to keep a small quantization error for large gradients. 

\begin{figure}[t]
	\centering
	\includegraphics[width=0.75\textwidth]{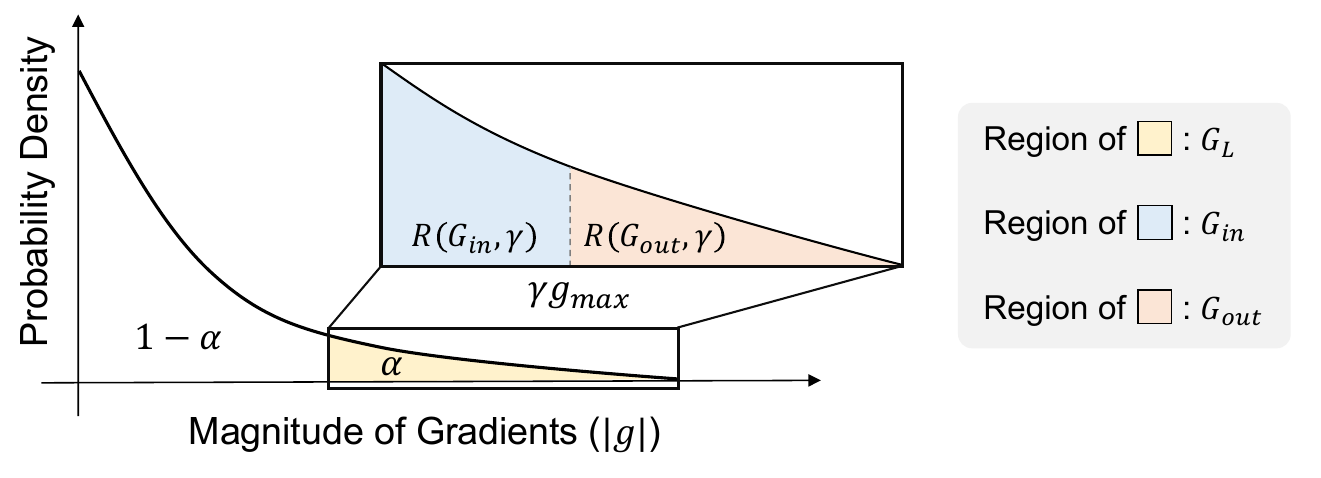}
	\vspace{-4mm}
	\caption{Probability density function (PDF) of gradient magnitudes for a single layer. The clip-in (blue) and clip-out (red) gradients,~$G_{in}$ and $G_{out}$, are subsets of large gradients~$G_L$~(yellow), and $G_{in}$ and $G_{out}$ are within and beyond the clipping value~$\gamma g_{max}$, respectively. See Sec.~\ref{sec:method_interval} for more details. (Best viewed in color.)}
	\vspace{-4mm}
	\label{fig:mag_distribute}
\end{figure}

\begin{figure}[t]
  \captionsetup[subfigure]{justification=centering}
  \captionsetup{font={small}}
  \begin{center}
     \begin{subfigure}{0.24\columnwidth}
        \centering
        \hspace{-0.2cm}\includegraphics[width=1\columnwidth]{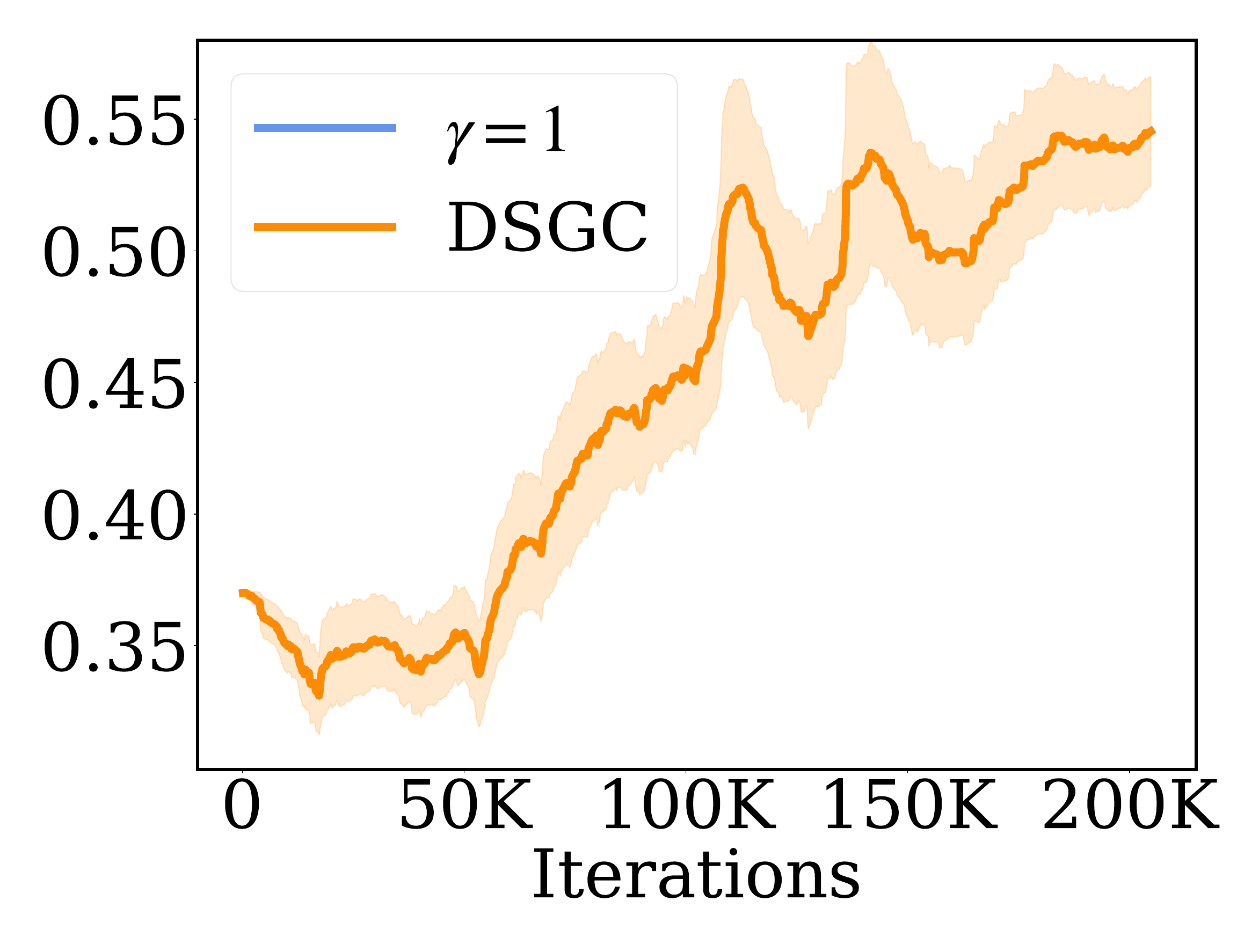}
        \caption{}
        \label{fig:2a}
     \end{subfigure}
     \begin{subfigure}{0.24\columnwidth}
        \centering
        \hspace{-0.4cm}\includegraphics[width=1.035\columnwidth]{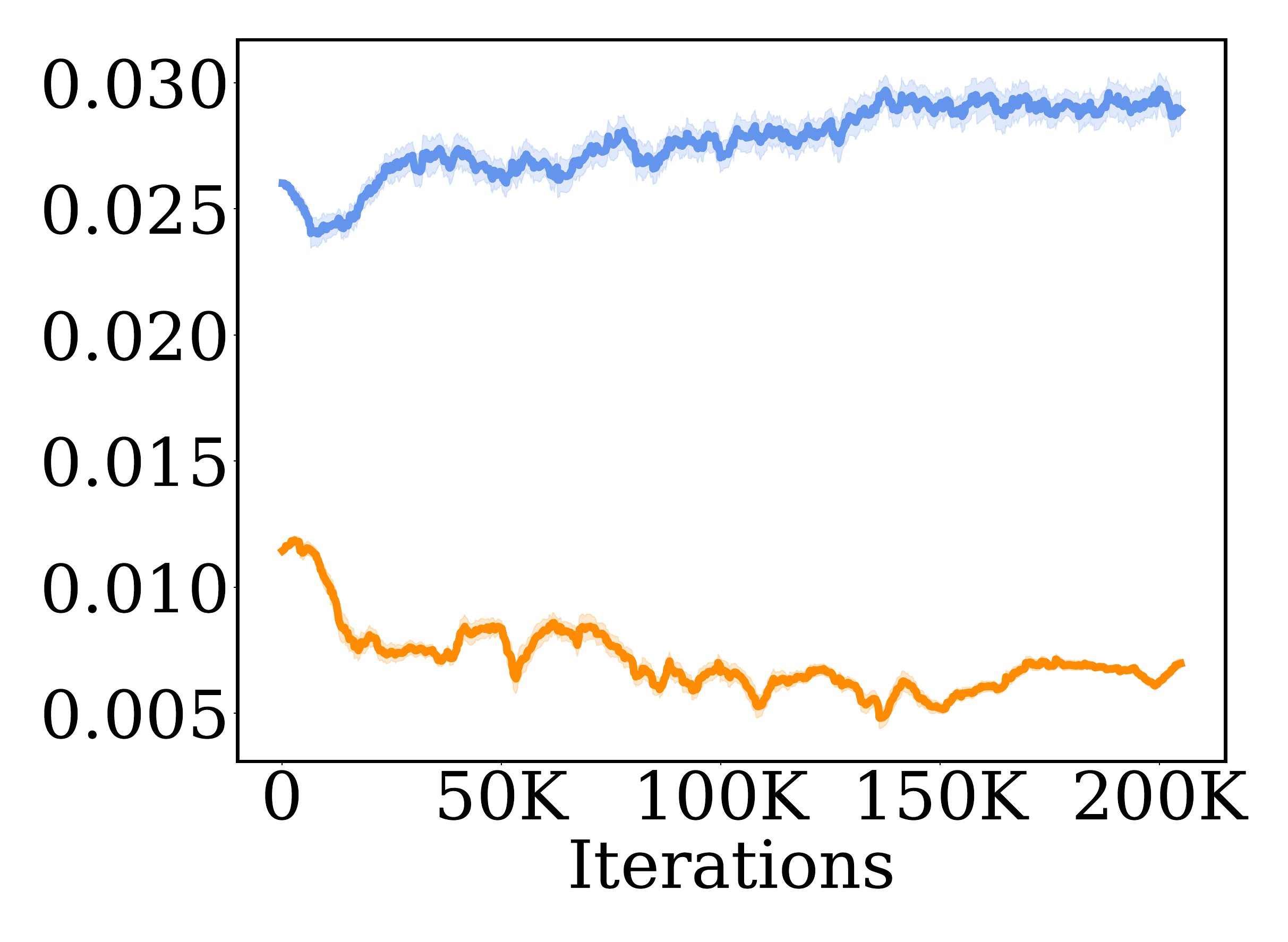}
        \caption{}
        \label{fig:2b}
     \end{subfigure}
     \begin{subfigure}{0.24\columnwidth}
        \centering
		\vspace{-0.05cm}
        \hspace{-0.2cm}\includegraphics[width=1\columnwidth]{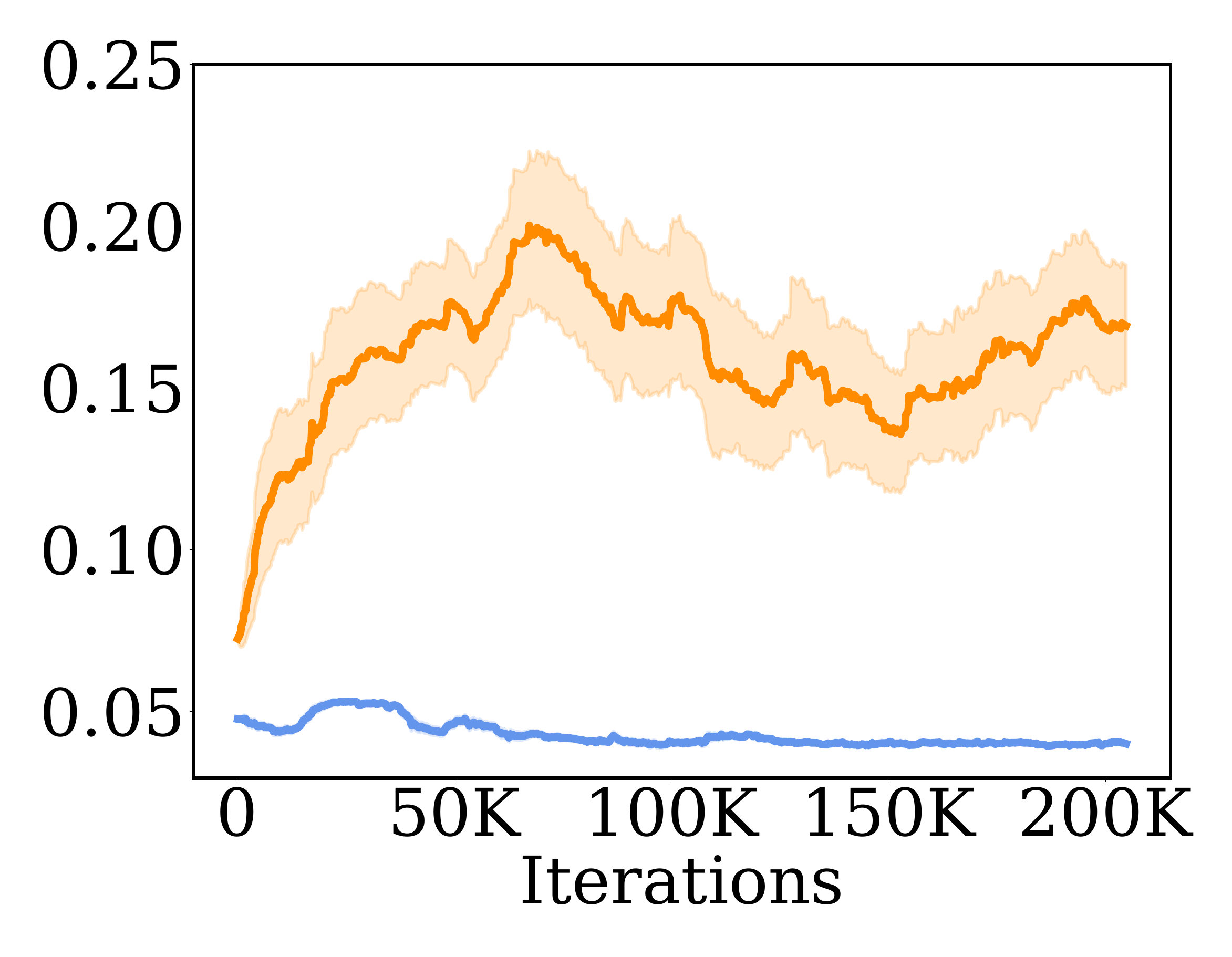}
        \caption{}
        \label{fig:2c}
     \end{subfigure}
     \begin{subfigure}{0.22\columnwidth}
        \centering
        \hspace{-0.1cm}\includegraphics[width=1\columnwidth]{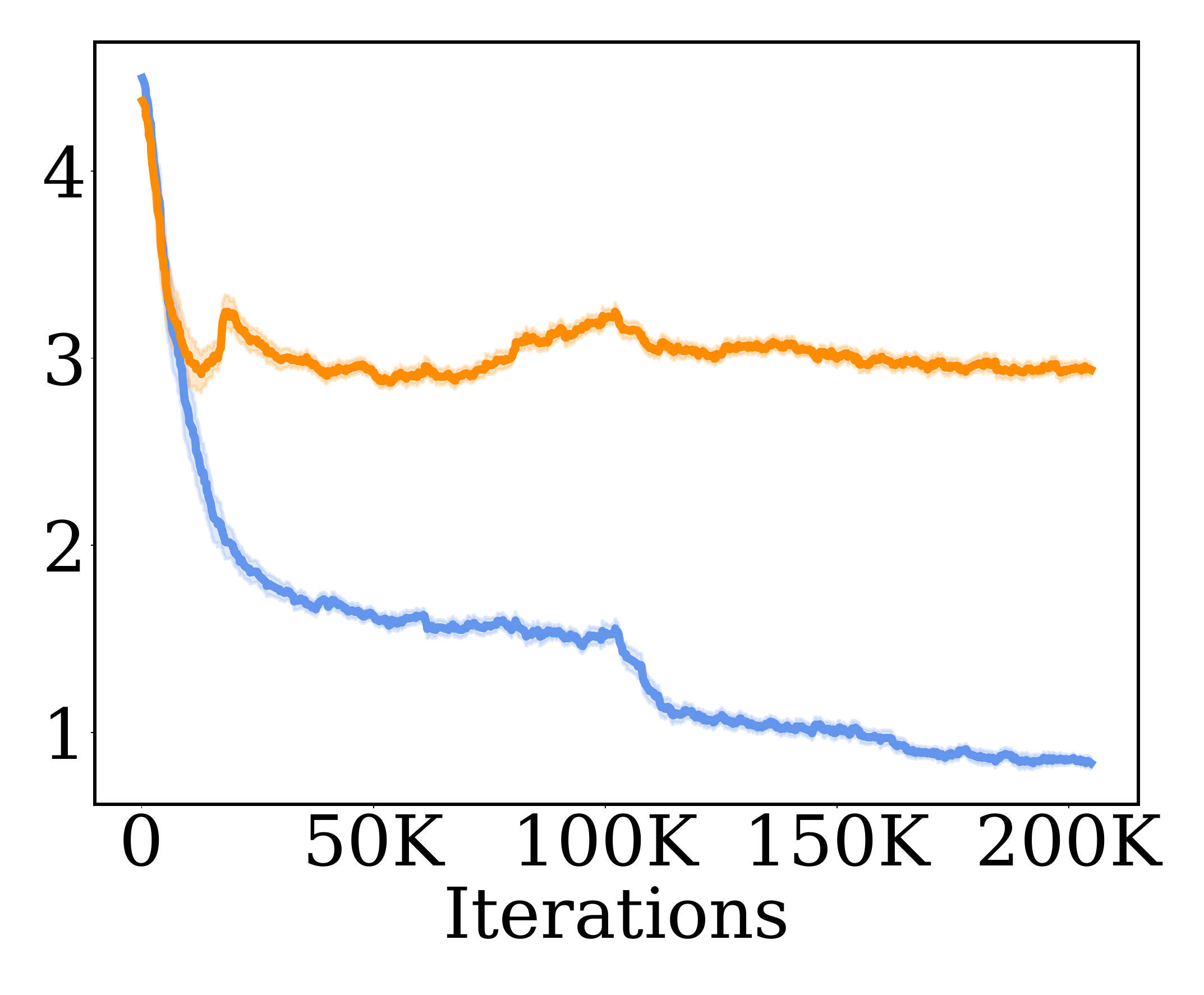}
         \caption{}
        \label{fig:2d}
     \end{subfigure}
  \end{center}
		\vspace{-0.6cm}
        \caption{Comparison of DSGC~\cite{zhu2020towards} and our baseline~($\gamma=1.0$). (a)~Clipping factor of~DSGC; (b)~The quantization error for entire gradients~$E(G)$; (c)~The quantization error for large gradients~$E(G_{L})$; (d)~Training loss. We visualize the quantization errors for entire gradients~$E(G)$ and large gradients~$E(G_{L})$, while tracking the clipping factor of DSGC in the 13th layer. Top-1 accuracies of DSGC and the baseline are 24.3 and 61.1, respectively, for the test split of CIFAR-100~\cite{krizhevsky2009learning}. (Best viewed in color.)}
        \vspace{-5mm}
     \label{fig:compare}
\end{figure}




\vspace{-0.2cm}
\subsection{Empirical analysis}
\vspace{-0.2cm}
\label{sec:analysis}
Here we present an analysis on how the quantization error for gradients affects the quantization performance. We train ResNet-20~\cite{he2016deep} on CIFAR-100~\cite{krizhevsky2009learning} using DSGC\footnote{Since the code for DSGC is not publicly available, we have reproduced it by ourselves.}~\cite{zhu2020towards} and our baseline in Sec.~\ref{sec:preliminary} with different clipping factors ($\gamma=0.4, 0.6, 0.8, 1.0$). We use 4-bit FXP values for each weight, activation, and gradient. We define an average quantization error for entire gradients, normalized w.r.t. the absolute maximum gradient~$g_{max}$ as follows:
  \begin{equation}
  \small
      \label{eq:lg_form}
      E({G}) = \frac{\sum_{g\in{G}}\vert g-Q(g)\vert}{N({G})g_{max}},
  \end{equation}
where~$N(G)$ counts the number of elements in the set~$G$. Similar to Eq.~\eqref{eq:lg_form}, we formulate the quantization error for large gradients as follows:
  \begin{equation}
  \small
      \label{eq:error_lg}
      E({G_L}) = \frac{\sum_{g\in{G_L}}\vert g-Q(g)\vert}{N({G_L})g_{max}}.
  \end{equation}
We define large gradients, denoted by~$G_L$, as a set of gradients whose magnitude is larger than a certain threshold splitting the density of gradients into $1-\alpha$ and $\alpha$~(Fig.~\ref{fig:mag_distribute}). That is, $\alpha$ is the ratio between the numbers of large gradients~$G_L$ and entire gradients~$G$,~\ie,~$\alpha=N(G_L)/N(G)$, which is a hyperparameter in our framework. We show an analysis on the quantization performance w.r.t. the quantization error for entire and large gradients, respectively, in Fig.~\ref{fig:compare}.
It provides a comparison of DSGC with a baseline ($\gamma=1$) in terms of the quantization error and accuracy. We can see from Figs.~\ref{fig:2a} and~\ref{fig:2b} that the clipping factors of DSGC are kept to small values, and the quantization error for entire gradients is smaller than that of the baseline. This enlarges the quantization error for large gradients significantly compared to the baseline (Fig.~\ref{fig:2c}). We can also see from Fig.~\ref{fig:2d} the training loss of DSGC having large error for large gradients increases in the middle of the training, \eg,~from the 50k-th to 100k-th iterations. Since large gradients mainly affect the training process~\cite{lee2021ewgs,ke2017lightgbm,han2017deep}, the quantization error for the large gradients deviates gradients significantly and causes unstable gradient flows, making the training unstable and subsequently degrading the quantization performance. We can conclude from Fig.~\ref{fig:compare} that lowering the quantization error for large gradients is more important than that for entire gradients in the low-bit FXP training.

\begin{figure*}[t]
  \captionsetup[subfigure]{justification=centering}
  \begin{center}
     \begin{subfigure}{0.24\columnwidth}
        \centering
        \includegraphics[width=1\columnwidth]{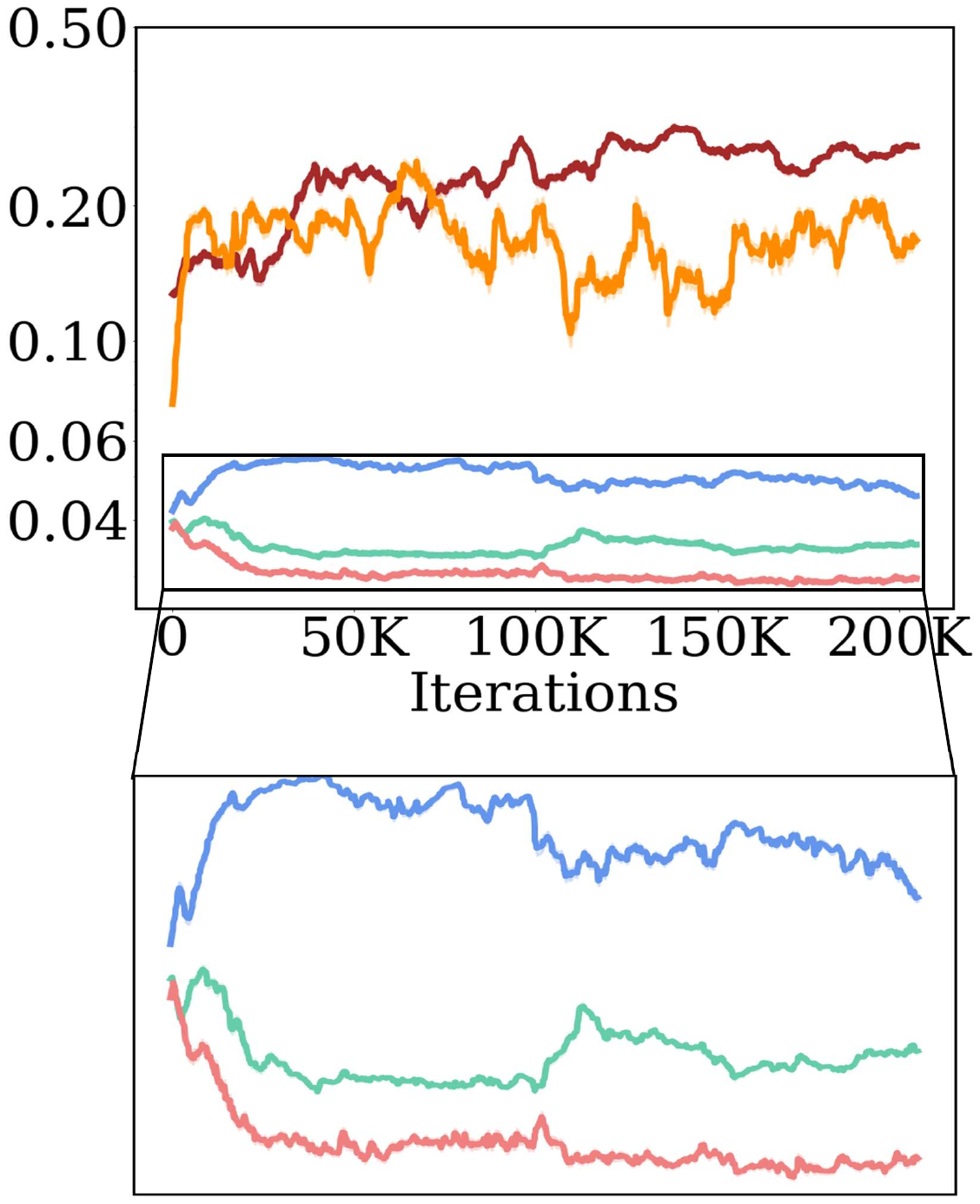}
		\vspace{-0.8cm}
        \caption{}
        \label{fig:3a}
     \end{subfigure}
     \begin{subfigure}{0.24\columnwidth}
        \centering
		\vspace{0.05cm}
        \includegraphics[width=1\columnwidth]{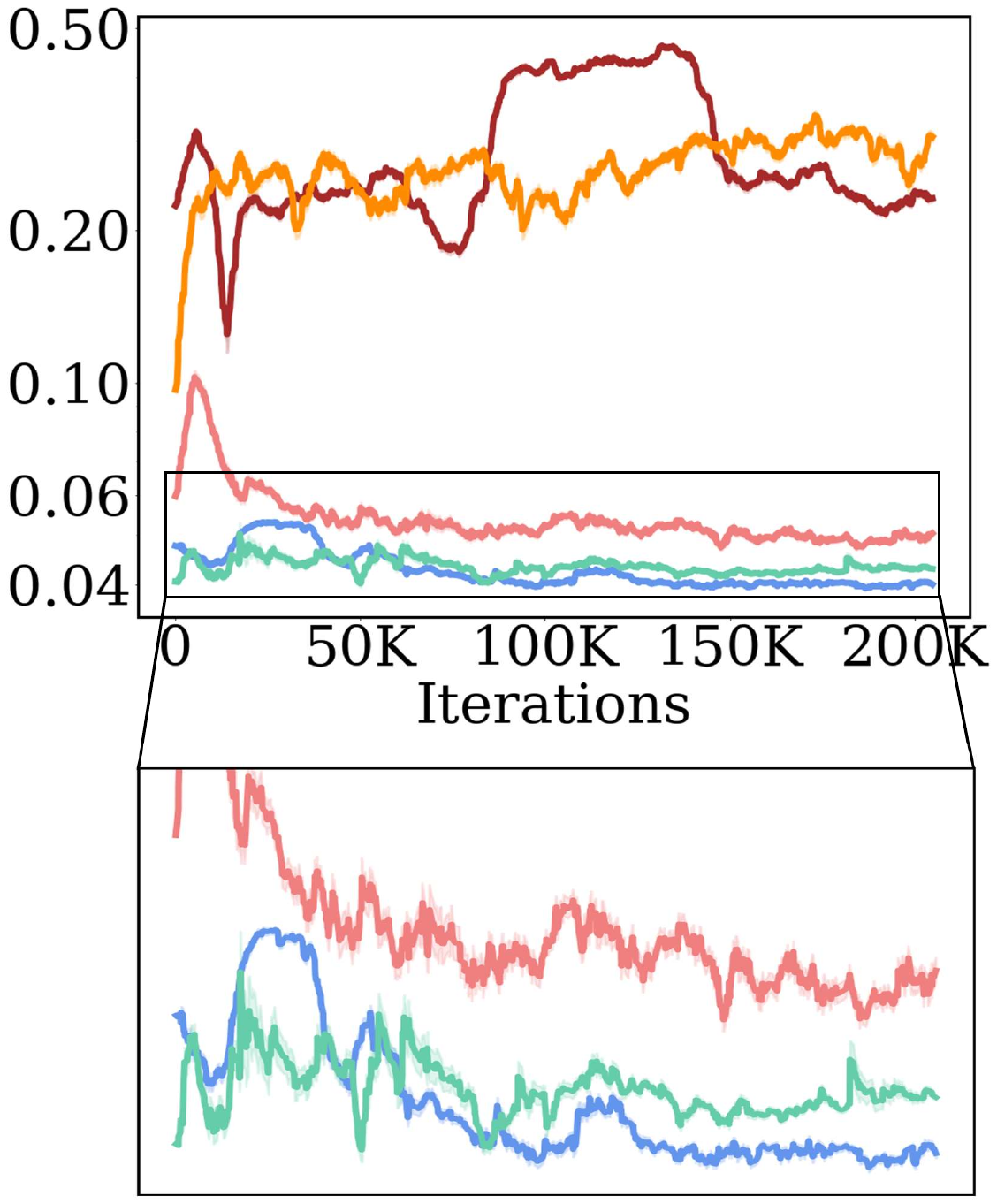}
		\vspace{-0.85cm}
        \caption{}
        \label{fig:3b}
     \end{subfigure}
     \begin{subfigure}{0.24\columnwidth}
        \centering
		\vspace{0.02cm}
        \includegraphics[width=1\columnwidth]{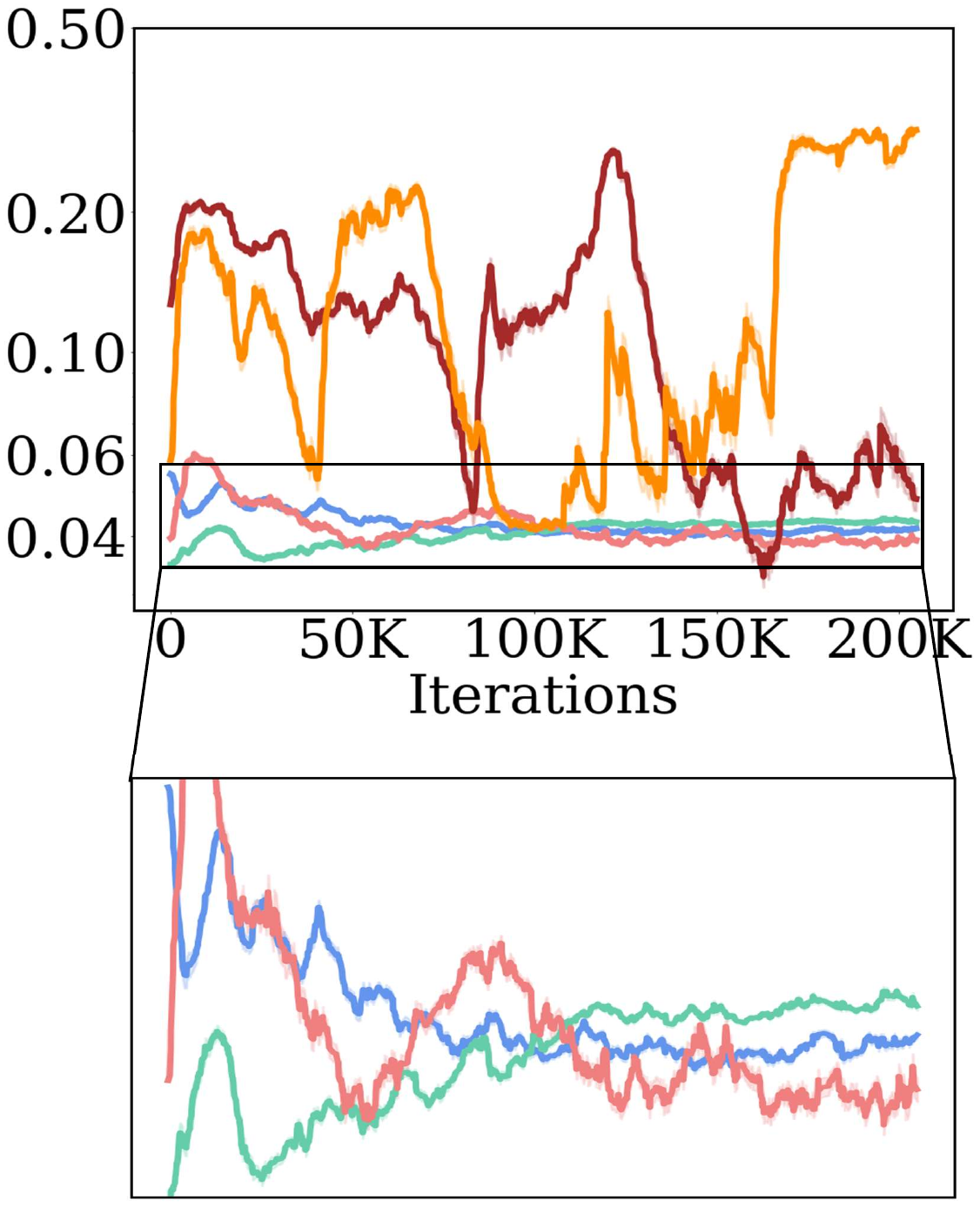}
		\vspace{-0.85cm}
        \caption{}
        \label{fig:3c}
     \end{subfigure}
     \begin{subfigure}{0.23\columnwidth}
        \centering
		\vspace{0.1cm}
        \includegraphics[width=1\columnwidth]{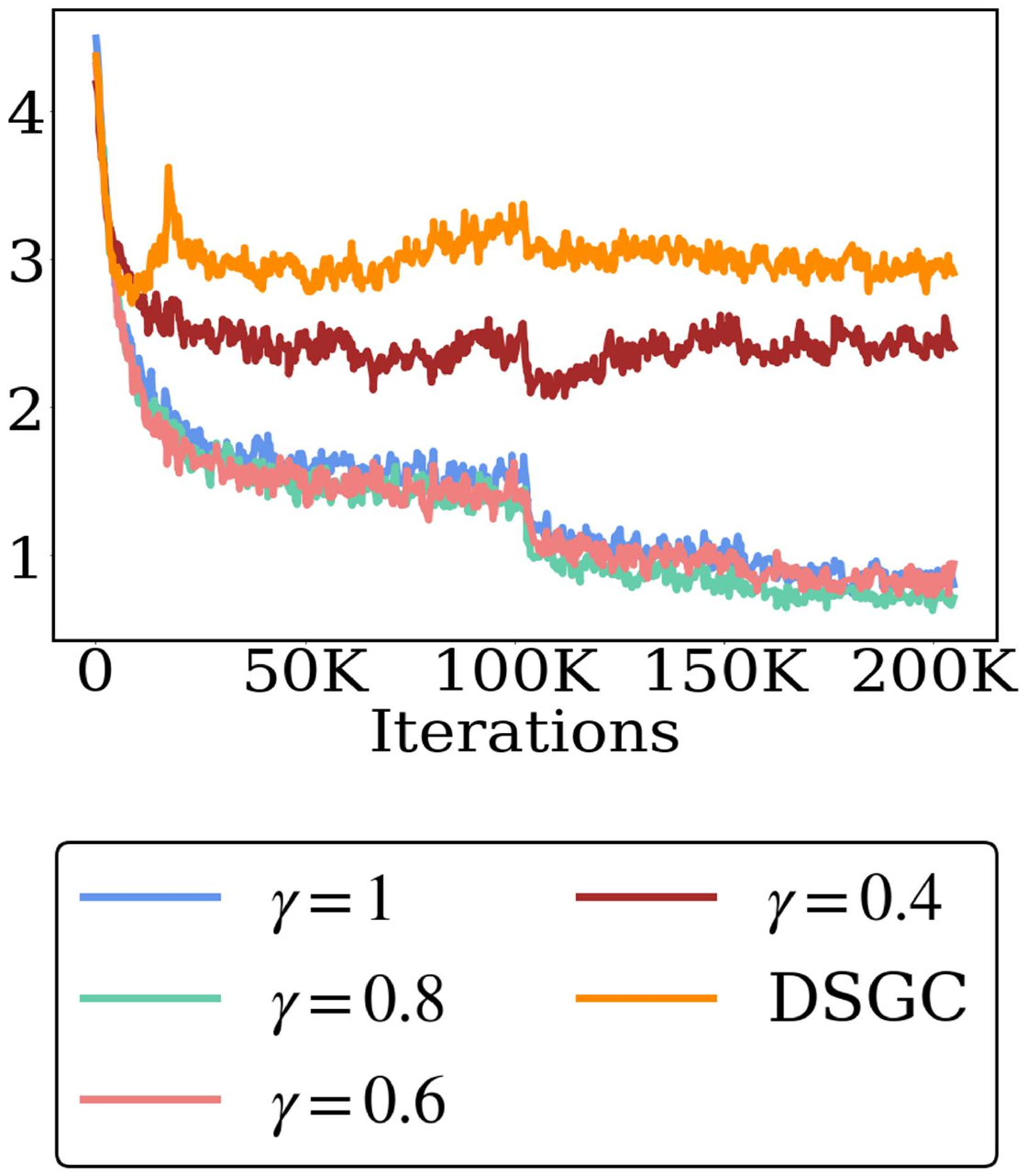}
		\vspace{-0.7cm}
        \caption{}
        \label{fig:3d}
     \end{subfigure}
  \end{center}
		\vspace{-0.7cm}  
        \caption{Empirical analysis on the quantization error for large gradients. (a-c)~$E(G_{L})$ in 13th, 15th, 17th layers, respectively; (d)~Training loss. Top-1 accuracies for the factors of 0.4, 0.6, and 0.8 are 30.3, 63.5, and 63.6, respectively, on the test split of CIFAR-100~\cite{krizhevsky2009learning}. (Best viewed in color.)} 
        \vspace{-5mm}
  \label{fig:empirical}
\end{figure*}


To delve deeper into this observation, we show in Fig.~\ref{fig:empirical} quantization errors for large gradients on different layers. 1) We can see that fixing a clipping factor to the small value (\ie, $\gamma=0.4$) brings the large quantization error for large gradients in 13, 15, 17th layers, similar to the observation from DSGC, and it shows worse quantization performance compared to other baselines ($\gamma=0.6,0.8,1.0$) providing smaller quantization errors for large gradients. This strengthens our motivation once more that lowering the quantization error for large gradients is a key factor to boost the performance in the FXP training. 2) We can see that a clipping factor lowering the quantization error for large gradients differs depending on the layer. For example, Figs.~\ref{fig:3a} and~\ref{fig:3b} show that fixing~$\gamma$ to 1 results in a larger quantization error for large gradients compared to others ($\gamma=0.6, 0.8$) in the 13th layer, while it shows a smaller error in the 15th layer. This is because distributions of gradients are different according to the layer~\cite{zhu2020towards,zhang2020fixed}. 3) Even in the same layer, a clipping factor lowering the quantization error for large gradients changes during training. For example, in the 17th layer (Fig.~\ref{fig:3c}), fixing~$\gamma$ to 0.6 leads to a larger quantization error for large gradients compared to other baselines in early iterations, while the error becomes smaller in later iterations.

Our empirical analysis suggests that lowering the quantization error for large gradients is better in terms of stability and accuracy in the FXP training, compared to lowering the error for entire gradients, even if the number of large gradients is very small compared to that of entire gradients~(Fig.~\ref{fig:compare}). It also indicates that we would adjust clipping factors adaptively for different layers and update them continually during training, in order to maintain a small quantization error for large gradients~(Fig.~\ref{fig:empirical}).

\vspace{-0.2cm}
\subsection{Interval update algorithm}
\vspace{-0.2cm}
\label{sec:method_interval}
We first derive an upper bound of the quantization error for large gradients (ULG), and obtain an optimal condition for the clipping factor~$\gamma$ lowering ULG. We then present our interval update algorithm that adjusts the clipping factor with a negligible computational overhead.


\vspace{-0.2cm}
\subsubsection{ULG.}
\label{sec:ulg_main}
We divide the large gradients into two parts, clip-in and clip-out gradients, denoted by~$G_{\mathrm{in}}$ and~$G_{\mathrm{out}}$, respectively. Specifically, the clip-in and clip-out gradients represent large gradients located inside and outside the quantization interval, respectively, \ie,~${G_{\mathrm{in}}}=\{g\vert \vert g\vert \leq \gamma g_{max},~g\in G_L\}$ and~${G_{\mathrm{out}}}=\{g\vert\vert g\vert > \gamma g_{max},~g\in G_L\}$~(Fig.~\ref{fig:mag_distribute}). The clip-in and clip-out gradients are hence influenced by the value of the clipping factor~$\gamma$. If we raise the clipping factor, the numbers of clip-in and clip-out gradients, $N(G_{\mathrm{in}}, \gamma)$ and $N(G_{\mathrm{out}}, \gamma)$, increase and decrease, respectively. The quantization error for large gradients in Eq.~\eqref{eq:error_lg} can be represented as follows: 
  \begin{equation}
  \small
      \label{eq:error_lg_detail}
      E({G_L}) = \frac{\sum_{g\in{G_{\mathrm{in}}}}\vert g-Q(g)\vert + \sum_{g\in{G_{\mathrm{out}}}}\vert g-Q(g)\vert}{N({G_L})g_{max}}.
  \end{equation}
Finding the clipping factor~$\gamma$ that minimizes~$E({G_L})$ is intractable. We instead derive ULG, and find the condition for the clipping factor lowering ULG. To this end, we define upper bounds of the quantization error within clip-in and clip-out gradients, separately: 1) The quantization error is maximized at the transition point, when the gradient exists inside the quantization interval,~\ie,~the clip-in gradient. In this case, the error quantifies the half of quantization step size, where the step size is $2\gamma g_{max}/(2^b - 2)$. The upper bound of quantization error for the clip-in gradient~$U_{\mathrm{in}}$ can then be set to~$\gamma g_{max}/(2^b - 2)$. 2) For the clip-out gradient, the quantized value is mapped to the end of the quantization interval,~\ie,~$Q(g)=\gamma g_{max}$ or $-\gamma g_{max}$. We thus compute the upper bound of error for the clip-out gradient~$U_{\mathrm{out}}$ as~$(1-\gamma) g_{max}$. Using the upper bounds of the error for clip-in and clip-out gradients,~$U_{\mathrm{in}}$ and~$U_{\mathrm{out}}$, we can derive ULG as follows (See the supplement for details):
  \begin{equation}
  \small
  \begin{aligned}
      \label{eq:quantizer}
      U({G_L}) &= \frac{U_{\mathrm{in}}N({G_{\mathrm{in}}}, \gamma) + U_{\mathrm{out}}N({G_{\mathrm{out}}}, \gamma)}{N({G_L})g_{max}}\\ 
	  				&= \left(\frac{\gamma}{2^b - 2}R({G_{\mathrm{in}}}, \gamma) + (1-\gamma)R({G_{\mathrm{out}}}, \gamma)\right)\frac{1}{\alpha},	  				
  \end{aligned}
  \end{equation}
where $R({G_{\mathrm{in}}}, \gamma)$ and $R({G_{\mathrm{out}}}, \gamma)$ are the ratios of clip-in and clip-out gradients, respectively, w.r.t. all gradients, defined as
  \begin{equation}
  \small  
      \label{eq:cratio}
      R({G_{\mathrm{in}}}, \gamma)=\frac{N({G_{\mathrm{in}}}, \gamma)}{N({G})},
      ~~R({G_{\mathrm{out}}}, \gamma)=\frac{N({G_{\mathrm{out}}}, \gamma)}{N({G})},
  \end{equation}
and
  \begin{equation}
  \small
  \label{eq:sum}
  \begin{aligned}
  	  \alpha = R({G_{\mathrm{in}}}, \gamma) + R({G_{\mathrm{out}}}, \gamma).
  \end{aligned}
  \end{equation}
 That is, $\alpha$ is the ratio between numbers of large gradients and entire gradients,~\ie,~$N(G_L)/N(G)$, which is a hyperparameter in our framework. Note that the union of clip-in and clip-out gradients,~$G_{\mathrm{in}}$ and~$G_{\mathrm{out}}$, is equal to the set of large gradients,~$G_L$~(Fig.~\ref{fig:mag_distribute}).

In order to find an optimal condition for the clipping factor, we take the derivative of ULG \wrt the factor~$\gamma$ as follows (See supplement for details):
  \begin{equation}
  \small
	\small
      \frac{dU({G_L})}{d\gamma} = \left(\frac{1}{2^b - 2}R({G_{\mathrm{in}}}, \gamma)-R({G_{\mathrm{out}}}, \gamma) + \left(1-\gamma-\frac{\gamma}{2^b - 2}\right)\frac{dR({G_{\mathrm{out}}}, \gamma)}{d\gamma}\right)\frac{1}{\alpha},
  \label{eq:diff}
  \end{equation}
Generally, the gradients follow a zero-centered distribution with a very long tail, but they are sparse around the tail~\cite{zhu2020towards}. Assuming that each side of the quantization interval exists around the tail, we can approximate that the change of the clip-out ratio~$R({G_{\mathrm{out}}})$ w.r.t. the clipping factor~$\gamma$ is negligible~(\ie,~$dR({G_{\mathrm{out}}}, \gamma)/d\gamma \approx 0$). Using the approximation, we represent Eq.~\eqref{eq:diff} as follows:
  \begin{equation}
  \small
  \label{eq:diff_appr}
  \begin{aligned}
      \frac{dU({G_L})}{d\gamma} \approx \left(\frac{1}{2^b - 2}R({G_{\mathrm{in}}}, \gamma)-R({G_{\mathrm{out}}}, \gamma)\right)\frac{1}{\alpha}.
  \end{aligned}
  \end{equation}
  Accordingly, the optimal clipping factor~$\gamma^*$ should satisfy the following condition:
  \begin{equation}
  \small
  \label{eq:opt_gamma}
	\frac{1}{2^b - 2}R({G_{\mathrm{in}}}, \gamma^*)-R({G_{\mathrm{out}}}, \gamma^*) = 0.
  \end{equation}
  The clipping factor~$\gamma^*$ satisfying the condition in Eq.~\eqref{eq:opt_gamma} can keep the small quantization error of large gradients for each layer and at every iteration.   
\vspace{-0.2cm}
\subsubsection{Updating clipping factors}
\label{sec:update}
Using Eqs.~\eqref{eq:sum} and~\eqref{eq:opt_gamma}, we can obtain the condition for optimal interval~$\gamma^*$ as follows:
  \begin{equation}
  \small
  \label{eq:optimal}
  \begin{aligned}
	R({G_{\mathrm{in}}}, \gamma^*) = \frac{2^b - 2}{2^b - 1} \alpha,~~R({G_{\mathrm{out}}}, \gamma^*) = \frac{1}{2^b - 1} \alpha.
  \end{aligned}
  \end{equation}
Manually searching the clipping factor that satisfies the condition in Eq.~\eqref{eq:optimal} is computationally demanding. We instead explore the relation between $R({G_{\mathrm{out}}}, \gamma)$ and the clipping factor~$\gamma$. Note that $R({G_{\mathrm{out}}}, \gamma)$ in Eq.~\eqref{eq:cratio} decreases slightly as the clipping factor~$\gamma$ increases, and vice versa~(Fig.~\ref{fig:mag_distribute}). Based on this, we design an algorithm that encourages the clipping factor to increase when the clip-out ratio is larger than the condition in Eq.~\eqref{eq:optimal}, \ie,~$R({G_{\mathrm{out}}}, \gamma) > \alpha/(2^b - 1)$, and vice versa. Concretely, we design the update scheme as follows:
  \begin{equation}
  \small
  \label{eq:itu}
  	  \gamma_i = \gamma_{i-1} + \beta \mathrm{sign}(T({G_{\mathrm{out}}}, \gamma_{i-1})),
  \end{equation}
where $\mathrm{sign}(\cdot)$ is a signum function and 
  \begin{equation}
  \small
  \label{eq:T}
	T({G_{\mathrm{out}}}, \gamma_{i-1}) = R({G_{\mathrm{out}}}, \gamma_{i-1}) - \frac{1}{2^b-1}\alpha,
  \end{equation}
which adjusts the direction of updating the clipping factor~$\gamma$. The scaling parameter~$\beta$(>0) controls the scale of $\mathrm{sign}(T({G_{\mathrm{out}}}, \gamma_{i-1}))$. If the clip-out ratio is larger than the condition,~\ie,~$R(G_{out},\gamma_{i-1})$ exceeds $\alpha$/$(2^b-1)$ corresponding to the condition in Eq.~\eqref{eq:optimal}, $T({G_{\mathrm{out}}}, \gamma_{i-1})$ is positive. The update algorithm thus raises the clipping factor of~$\gamma_{i-1}$ in the~$(i-1)$th iteration. Conversely, the algorithm reduces the clipping factor by $\beta$ when~$R(G_{out},\gamma_{i-1})$ is below the condition in Eq.~\eqref{eq:optimal}. Accordingly, the clipping factor is adjusted adaptively to satisfy the condition in Eq.~\eqref{eq:optimal}, maintaining a small quantization error for large gradients. We provide algorithm table describing the process of updating the clipping factor in the supplementary material.


\section{Experiments}
\vspace{-0.3cm}
\subsection{Experimental details}
\vspace{-0.3cm}
\label{sec:detail}

\subsubsection{Image classification.} We quantize weights, activations, and gradients for a family of ResNets~\cite{he2016deep} and MobileNetV2~\cite{sandler2018mobilenetv2} on CIFAR-100~\cite{krizhevsky2009learning} and ImageNet~\cite{deng2009imagenet}. Following the works of~\cite{zhu2020towards,zhou2016dorefa,sun2019hybrid,das2018mixed,sun2020ultra}, we do not quantize the first and last layers, and use the stochastic rounding technique~\cite{gupta2015deep} for gradient quantization. We use the Adam optimizer~\cite{kingma2014adam} with a learning rate of 1e-5 for all networks to train clipping values for weights and activations, where they are initialized with the approach of~\cite{lee2021ewgs}. Note that we do not learn the clipping value for gradient quantization. We train network weights from scratch with random initialization using the SGD optimizer, where the initial learning rates are set to 1e-1 and 5e-2 for ResNets and MobileNetV2, respectively. For ResNet-20, we train quantized networks for 160 epochs on CIFAR-100 with a batch size of 128, and a weight decay of 1e-4. We use a step learning rate schedule with a decay of 0.1 at epoch 80 and 120. For the ResNet-18, -34, and -50 architectures, quantized networks are trained on ImageNet for 100 epochs with a batch size of 256 and the weight decay of 1e-4. We adopt a step learning rate schedule with a decay of 0.1 at epoch 30, 60, and 90. For MobileNetV2, we train quantized networks for 150 epochs on ImageNet with a batch size of 256 and a weight decay of 4e-5. We use a cosine annealing technique~\cite{loshchilov2016sgdr} for learning rate decay. Following~\cite{sun2020ultra,sun2019hybrid}, we do not quantize the depth-wise convolutional layers in MobileNetV2. We fix the scaling parameter~$\beta$ to 1e-3 for all experiments. We set the ratio~$\alpha$ as a hyperparameter of~$\tau$ divided by the total number of gradients in the network. We find $\tau$ by a grid search with ResNet-20~\cite{he2016deep} on CIFAR-100~\cite{krizhevsky2009learning}, and fix it for all experiments~\footnote{We provide an analysis on the hyperparmeters in the supplementary material.}.

\vspace{-0.5cm}
\subsubsection{Object detection.} We quantize Faster R-CNN~\cite{ren2015faster} exploiting the ResNet-50 architecture as a backbone. We use the SGD optimizer with an initial learning rate of 1e-2, a weight decay of 1e-4, and a batch size of 16. We train the model for 90k iterations with a step learning rate scheduler on the COCO dataset~\cite{lin2014microsoft}, where the learning rate is reduced by a factor of 0.1 at 60k and 80k iterations.
\vspace{-0.5cm}
\subsubsection{Super-resolution.} To demonstrate the generalizability of our method, we apply our method to image super-resolution. To this end, we quantize weights, activations, and gradients for EDSR~\cite{lim2017enhanced} on the DIV2K dataset~\cite{agustsson2017ntire}, and train the model for 300 epochs using the Adam optimizer with a batch size of 16. The learning rate is initialized with 2e-4, and we decay the learning rate by a factor of 0.5 every 100 epochs.

\begin{table*}[t] 
  \setlength{\tabcolsep}{0.3em}
  \small
  \centering
  \bgroup
  \def\arraystretch{1.1}
  \caption{Quantitative comparison of gradient quantization methods on image classification. We report results on the validation split of ImageNet~\cite{deng2009imagenet} in terms of a top-1 accuracy. W/A/G: Bit-precision of weights/activations/gradients; FP: Results obtained by full-precision models; $\dagger$: Results reproduced by ourselves. Numbers in bold and parentheses are the best result and accuracy improvements or degradations, w.r.t full-precision models, respectively.}
  \vspace{-0.3cm}
  \label{tab:large_data}
  \begin{adjustbox}{width=0.9\columnwidth,center}
  \begin{tabular}{C{1.9cm}lC{0.9cm}C{2.2cm}C{2.2cm}C{2.2cm}C{2.2cm}}
  \hline
  \multicolumn{1}{c}{\multirow{2}{*}{Architecture}} & \multicolumn{1}{c}{\multirow{2}{*}{Method}}& \multicolumn{5}{c}{Bit-width (W/A/G) } \\

 & & FP & 8/8/8 & 6/6/6 & 5/5/5 & 4/4/4\\  
  \hline
    
  \multicolumn{1}{c}{\multirow{9}{*}{\shortstack[c]{ResNet-18}}} & Baseline ($\gamma$=1.0)       		& 70.4 & 70.3 ($-$0.1) & 69.4 ($-$1.0) & 67.2 ($-$3.2) & 63.3 ($-$7.1) \\
  & Baseline ($\gamma$=0.8) 				& 70.4 & 69.6 ($-$0.8) & 68.7 ($-$1.7) & 67.3 ($-$3.1) & 63.9 ($-$6.5) \\
  & Baseline ($\gamma$=0.6) 				& 70.4 & 63.0 ($-$7.4) & 61.7 ($-$8.7) & \hspace{1mm} 59.6 ($-$10.8) & \hspace{1mm}50.2 ($-$20.2) \\
  & WAGEUBN~\cite{yang2020training}			& 70.3 & 66.9 ($-$3.4) & - & - & -\\ 
  & DSGC~\cite{zhu2020towards}			& 70.3 & 69.7 ($-$0.6) & - & - & -\\ 
  & DSGC$^\dagger$~\cite{zhu2020towards}	& 70.4 & 69.7 ($-$0.7) & \hspace{1mm} 55.8 ($-$14.6) & \hspace{1mm} 57.2 ($-$13.2) & \hspace{1mm}29.8 ($-$40.6)\\ 
  & IQB~\cite{liu2022IQB}			& 70.1 & 69.7 ($-$0.4) & - & - & - \\   
  & IQB$^\dagger$~\cite{liu2022IQB}			& 70.4 & 70.2 ($-$0.2) & 69.5 ($-$0.9) & 67.9 ($-$2.5) & ~59.9 ($-$10.5)\\ 
  & \cellcolor[HTML]{DAE8FC}Ours									& \cellcolor[HTML]{DAE8FC}70.4 & \cellcolor[HTML]{DAE8FC}\bf{70.4 ($-$0.0)} & \cellcolor[HTML]{DAE8FC}\bf{70.0 ($-$0.4)} & \cellcolor[HTML]{DAE8FC}\hspace{1mm} \bf{69.0} ($-$1.4) & \cellcolor[HTML]{DAE8FC}\bf{67.1 ($-$3.3)}\\ 
  
  \hline
  
  \multicolumn{1}{c}{\multirow{6}{*}{\shortstack[c]{ResNet-34}}}   
  & Baseline ($\gamma$=1.0)       		& 73.9 & 73.3 ($-$0.6) & 72.1 ($-$1.8) & \hspace{1mm} 69.7 ($-$4.2) & 62.9 (-11.0)\\
  & WAGEUBN~\cite{yang2020training}			& 73.7 & 68.5 ($-$5.2) & - & - & -\\ 
  & DSGC~\cite{zhu2020towards}			& 73.7 & 73.3 ($-$0.4) & - & - & -\\ 
  & DSGC$^\dagger$~\cite{zhu2020towards}	& 73.9 & 73.1 ($-$0.8) & \hspace{1mm} 62.4 ($-$11.5) & \hspace{1mm} 61.2 ($-$12.7) & 31.8 ($-$42.1)\\ 
  & IQB$^\dagger$~\cite{liu2022IQB}	& 73.9 & 73.3 ($-$0.6) & 72.4 ($-$1.5) & 70.9 ($-$3.0) & 60.1 ($-$13.8) \\ 
  & \cellcolor[HTML]{DAE8FC}Ours									& \cellcolor[HTML]{DAE8FC}73.9 & \cellcolor[HTML]{DAE8FC}{\bf{73.7}} \bf{($-$0.2)} & \cellcolor[HTML]{DAE8FC}\bf{72.7 ($-$1.2)} & \cellcolor[HTML]{DAE8FC}\hspace{1mm} \bf{71.9} ($-$2.0) & \cellcolor[HTML]{DAE8FC}\bf{69.2 ($-$4.7)}\\ 
  
  \hline
  
  \multicolumn{1}{c}{\multirow{4}{*}{\shortstack[c]{ResNet-50}}}   
  & Baseline ($\gamma$=1.0)       		& 76.2 & 75.9 ($-$0.3) & 73.7 ($-$2.5) & 69.5 ($-$6.7) & -\\
  & WAGEUBN~\cite{yang2020training}			& 76.6 & 69.1 ($-$7.5) & - & - & -\\ 
  & DSGC~\cite{zhu2020towards}			& 76.6 & {\bf{76.3}} ($-$0.3) & - & - & -\\ 
  & \cellcolor[HTML]{DAE8FC}Ours									& \cellcolor[HTML]{DAE8FC}76.2 & \cellcolor[HTML]{DAE8FC}76.0 {\bf{($-$0.2)}} & \cellcolor[HTML]{DAE8FC}\bf{74.2 ($-$2.0)} & \cellcolor[HTML]{DAE8FC}\bf{72.2} ($-$4.0) & \cellcolor[HTML]{DAE8FC}-\\ 
  
  \hline
  
  \multicolumn{1}{c}{\multirow{3}{*}{\shortstack[c]{MobileNetV2}}}   
  & Baseline ($\gamma$=1.0)       		& 71.9 & 71.3 ($-$0.6) & 70.9 ($-$1.0) & 68.0 ($-$3.9) & -\\
  & DSGC~\cite{zhu2020towards}			& 72.4 & 71.2 ($-$1.2) & - & - & -\\ 
  & \cellcolor[HTML]{DAE8FC}Ours									& \cellcolor[HTML]{DAE8FC}71.9 & \cellcolor[HTML]{DAE8FC}\bf{71.6} \bf{($-$0.3)} & \cellcolor[HTML]{DAE8FC}\bf{71.0} ($-$0.9) & \cellcolor[HTML]{DAE8FC}\bf{69.0} ($-$2.9) & \cellcolor[HTML]{DAE8FC}-\\ 
  
  \hline
  
\hline
  \end{tabular}
  \end{adjustbox}
  \egroup
  \vspace{-0.3cm}
\end{table*}

\begin{table*}[t!] 
  \setlength{\tabcolsep}{0.3em}
  \small
  \centering
  \bgroup
  \def\arraystretch{1.15}
  \caption{Quantitative comparison of gradient quantization methods on image classification. We report results on the test split of CIFAR-100~\cite{krizhevsky2009learning} in terms of a top-1 accuracy.}
  \label{tab:small_data}
  \begin{adjustbox}{width=0.7\columnwidth,center}
  \begin{tabular}{C{1.9cm}lC{0.9cm}C{2.1cm}C{2.1cm}}
  \hline
  \multicolumn{1}{c}{\multirow{2}{*}{Architecture}} & \multicolumn{1}{c}{\multirow{2}{*}{Method}}& \multicolumn{3}{c}{Bit-width (W/A/G) } \\

    & & FP & 8/8/8 & 4/4/4\\  
  \hline
  \multicolumn{1}{c}{\multirow{5}{*}{\shortstack[c]{ResNet-20}}} & Baseline ($\gamma$=1.0)       		& 66.9 & 66.1 ($-$0.8) & 61.1 ($-$5.8)\\
  & Baseline ($\gamma$=0.8) 				& 66.9 & 66.2 ($-$0.7) & 63.6 ($-$3.3)\\
  & Baseline ($\gamma$=0.6) 				& 66.9 & 65.7 ($-$1.2) & 63.5 ($-$3.4)\\
  & DSGC$^\dagger$~\cite{zhu2020towards}	& 66.9 & 66.1 ($-$0.8) & \hspace{0.5mm} 24.3 ($-$42.6)\\ 
  & \cellcolor[HTML]{DAE8FC}Ours									& \cellcolor[HTML]{DAE8FC}66.9 & \cellcolor[HTML]{DAE8FC}\bf{66.8 ($-$0.1)} & \cellcolor[HTML]{DAE8FC}\hspace{0.1mm} \bf{65.0 ($-$1.9)}\\ 
  \hline
   
  \end{tabular}
  \end{adjustbox}
  \egroup
  \vspace{-6mm}
\end{table*}

\vspace{-0.4cm}
\subsection{Results}
\vspace{-0.3cm}
\label{sec:results}
\subsubsection{Image classification.} We apply our quantization method to various CNN architectures, including a family of ResNets~\cite{he2016deep}, MobileNetV2~\cite{sandler2018mobilenetv2}. We compare our approach with state-of-the-art FXP training methods\footnote{For a fair comparison, we compare our approach with FXP training methods exploiting layer-wise quantization, and do not perform the comparisons with FLP training methods, sample-wise quantization, and channel-wise quantization.}~\cite{yang2020training,zhu2020towards} on ImageNet and CIFAR-100 in Tables~\ref{tab:large_data} and~\ref{tab:small_data}, respectively. All numbers except for the baselines, DSGC$^\dagger$, and IQB$^\dagger$\footnote{Since the code for IQB is not publicly available, we reproduce the piecewise FXP format, and apply it to quantize the gradients in our baseline.} are taken from the work of~\cite{zhu2020towards} including the results of full-precision models. Results for the transformer architecture can be found in the supplementary material.
From these tables, we observe four things: 1)~Our method outperforms other FXP training methods by a significant margin in terms of a top-1 accuracy, regardless of datasets, network architectures, and bit-widths. The accuracy of DSGC is slightly better than ours for the 8/8/8-bit setting only on the ResNet-50 architecture in Table~\ref{tab:large_data}. Nevertheless, ours shows a lower accuracy drop w.r.t the full-precision model. Note that the full-precision model in DSGC also shows a higher accuracy, possibly due to different training settings for,~\eg,~the number of epochs and learning rate scheduling. 2)~We can see that the accuracy drop of DSGC becomes severe as bit-widths decrease. A plausible reason is that reducing the bit-width increases the quantization error for entire gradients, and the quantization interval of DSGC becomes narrower in order for keeping a small error for entire gradients. It incurs a significant quantization error for large gradients, and the performance in turn degrades drastically. Compared to DSGC, our method provides better results consistently, confirming once more that lowering the quantization error for large gradients is important in the FXP training. 3)~Our method shows better results compared to the state of the art, including DSGC and IQB, in particularly low-bit settings (\ie,~6/6/6, 5/5/5, and 4/4/4-bit settings). For example, our method performs better than IQB~\cite{liu2022IQB} employing a piecewise FXP format for gradient quantization, when training ResNet-18 and -34 in 4/4/4 and 5/5/5-bit settings, and obtains the superior results over the baseline when training in 4/4/4 and 5/5/5-bit settings. This suggests that maintaining a small error for large gradients is effective to improve the quantization performance in the low-bit settings. 4) We can clearly observe that ours gives better results than the baselines with various architectures consistently, especially in the 4/4/4 and 5/5/5-bit settings. This indicates that maintaining a small quantization error for large gradients, regardless of the layers or training iterations, is significant in the FXP training.

\begin{figure*}[t]
  \captionsetup[subfigure]{justification=centering}

  \begin{center}
     \begin{subfigure}{0.24\columnwidth}
        \centering
        \hspace{-0.2cm}\includegraphics[width=1\columnwidth]{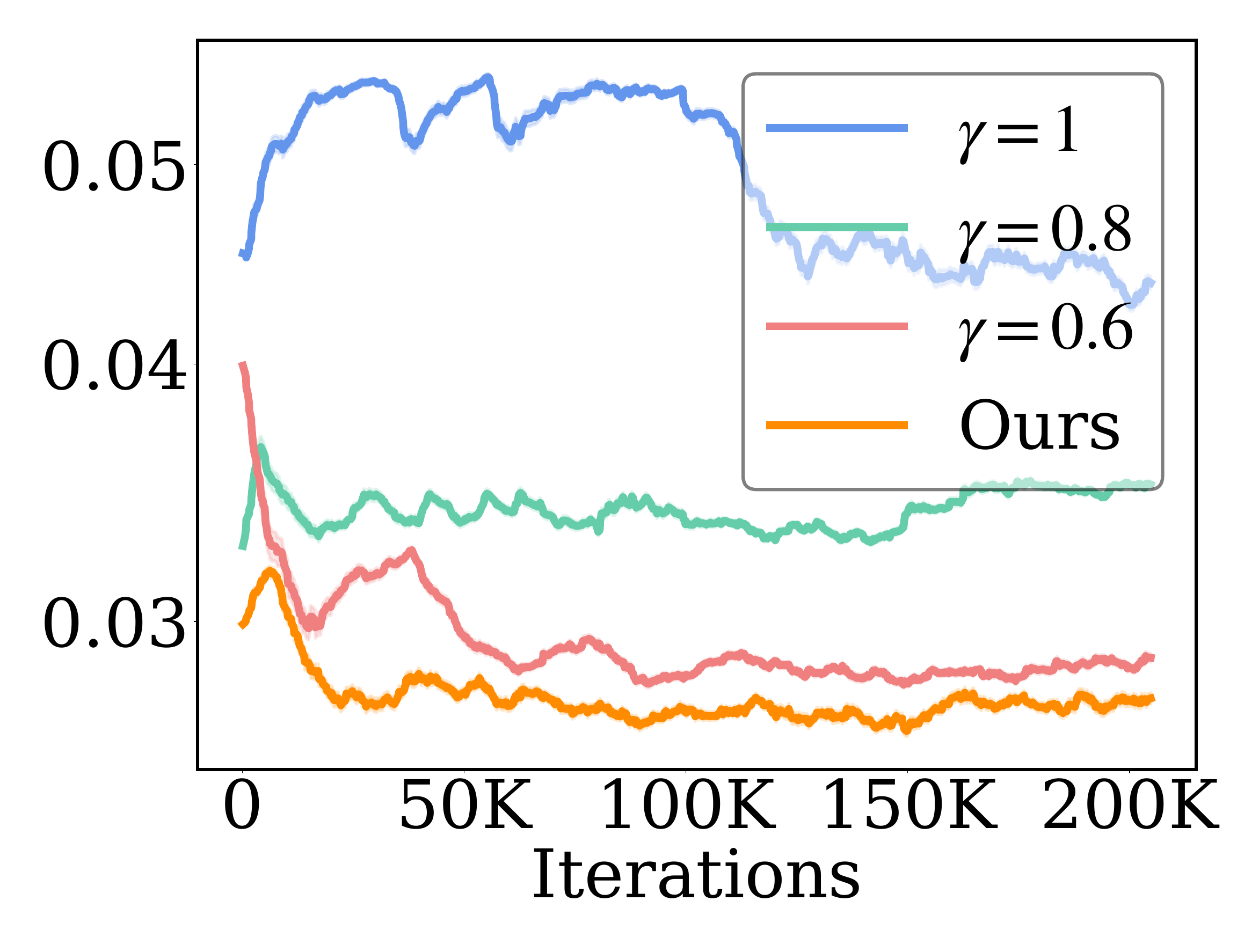}
        \vspace{-0.1cm}
        \caption{}
        \label{fig:4a}
     \end{subfigure}
     \begin{subfigure}{0.24\columnwidth}
        \centering
        \hspace{-0.2cm}\includegraphics[width=1\columnwidth]{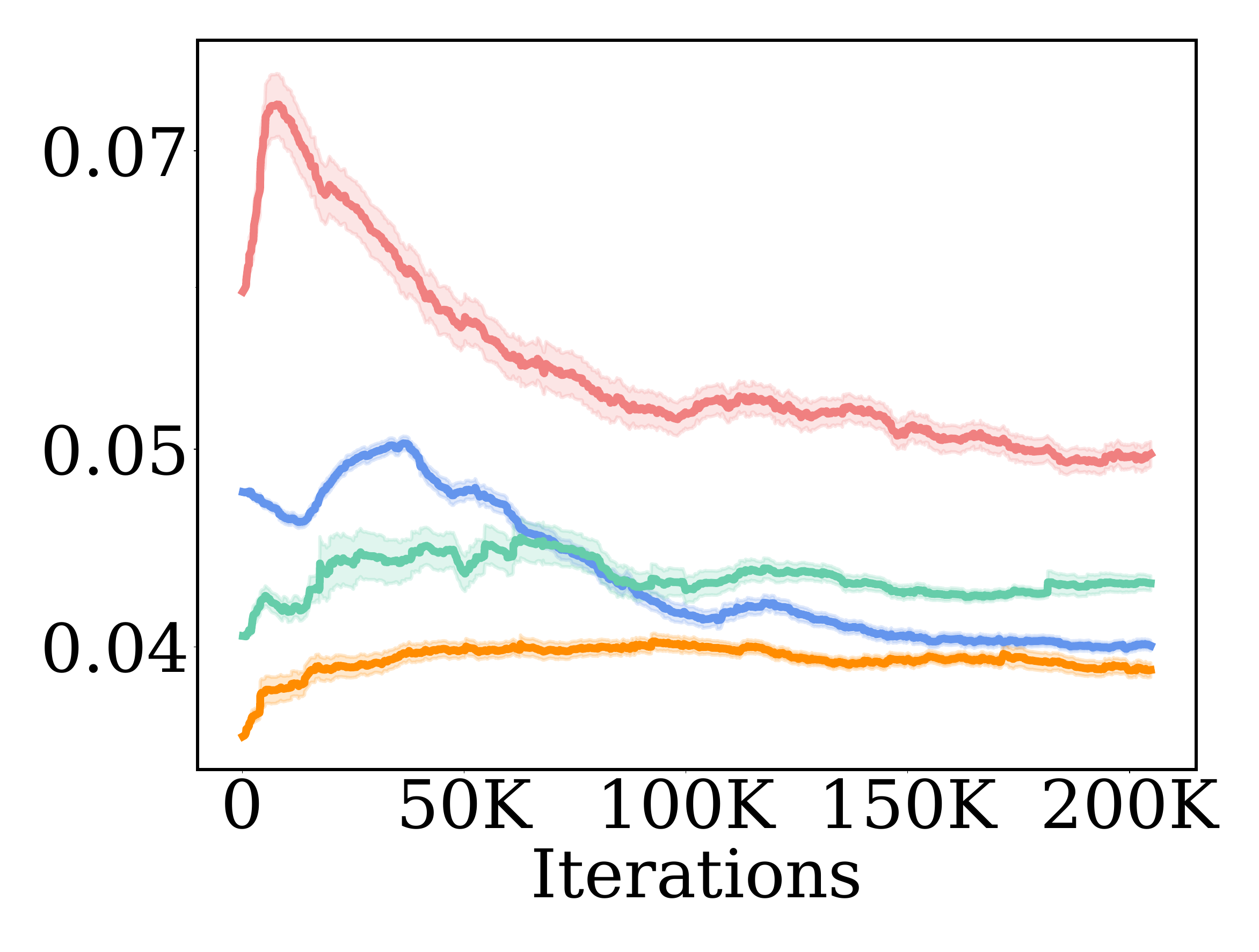}
        \vspace{-0.1cm}
        \caption{}
        \label{fig:4b}
     \end{subfigure}
     \begin{subfigure}{0.24\columnwidth}
        \centering
        \hspace{-0.2cm}\includegraphics[width=1\columnwidth]{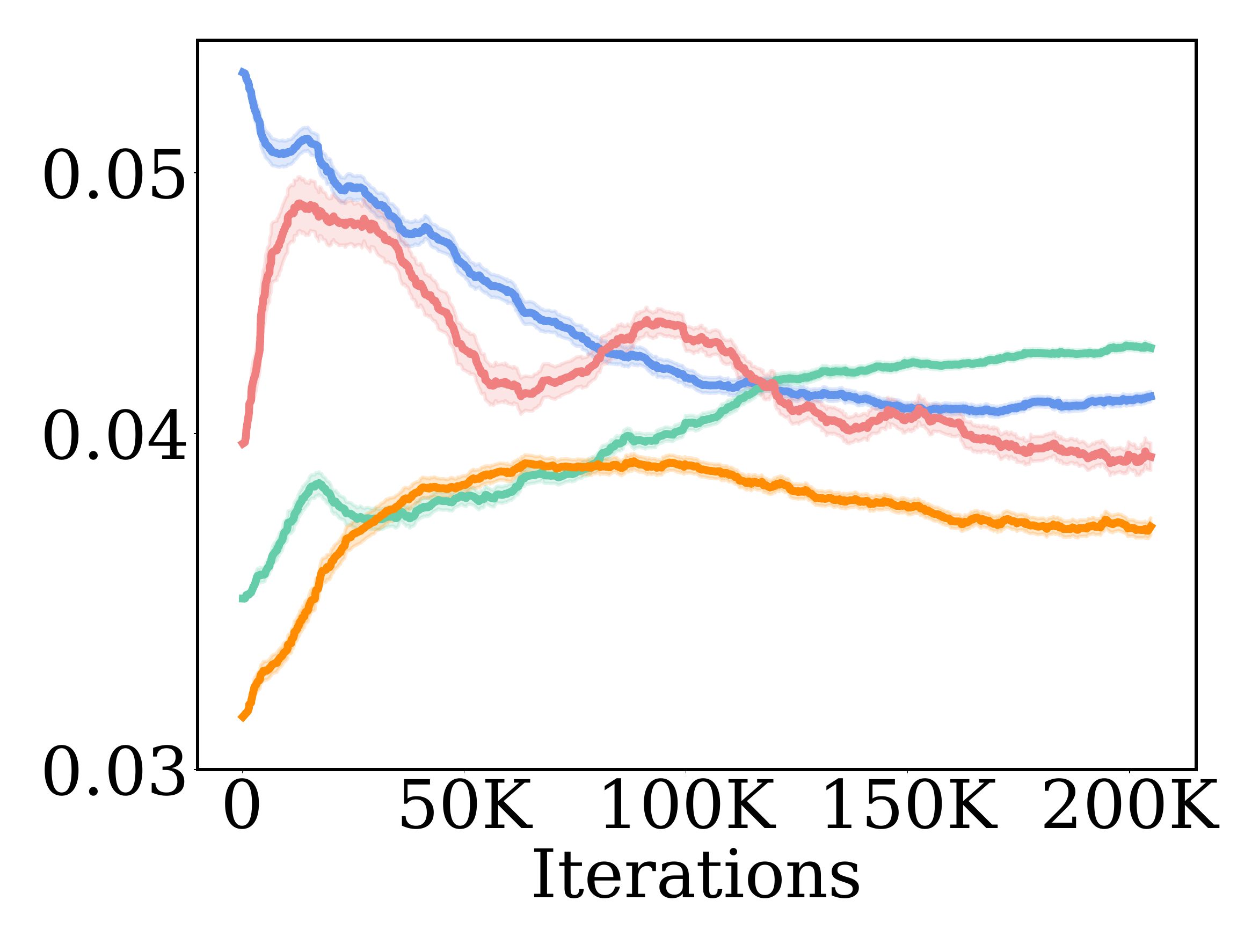}
        \vspace{-0.1cm}
        \caption{}
        \label{fig:4c}
     \end{subfigure}
     \begin{subfigure}{0.23\columnwidth}
        \centering
        \hspace{-0.2cm}\includegraphics[width=1\columnwidth]{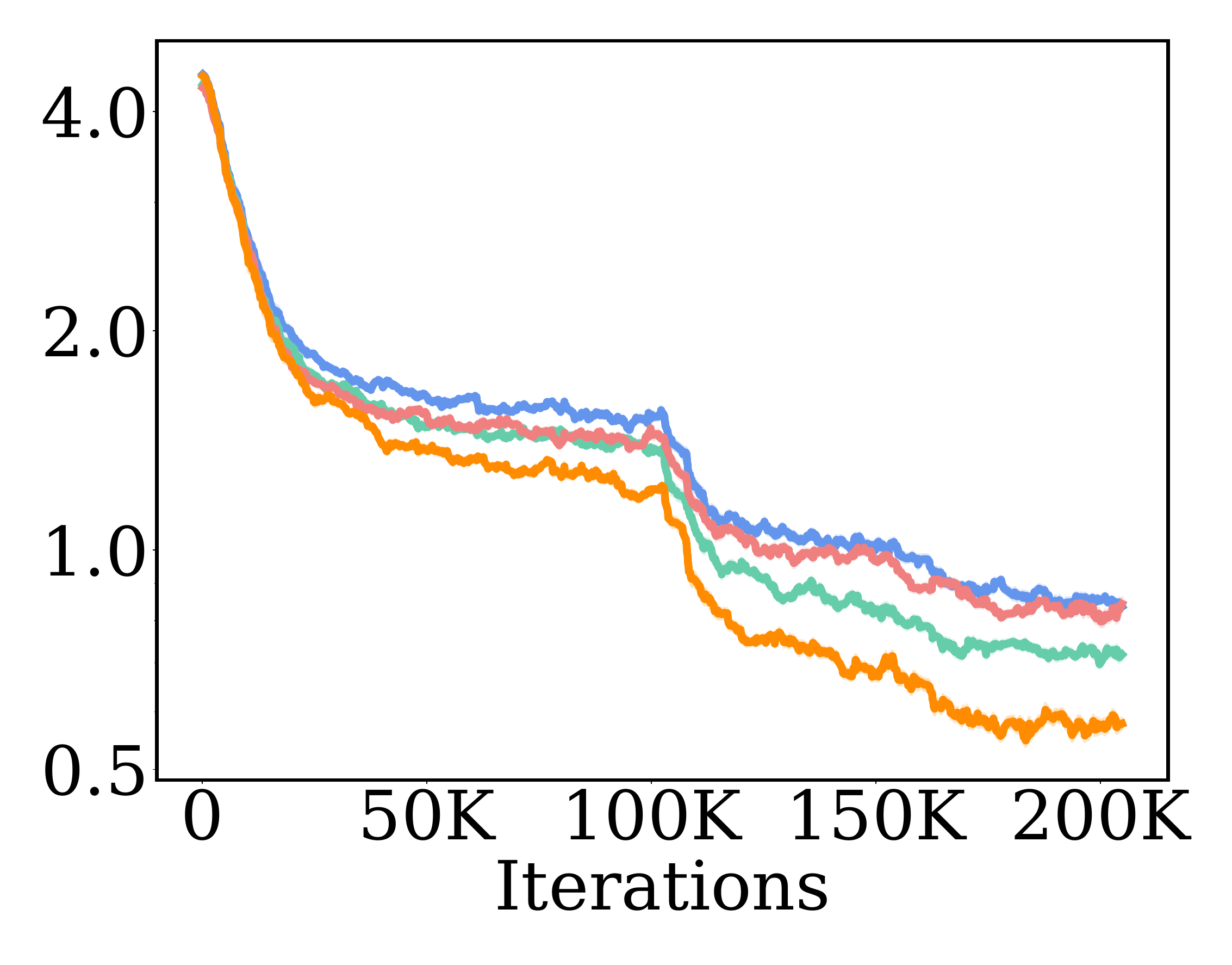}
        \vspace{-0.1cm}
        \caption{}
        \label{fig:4d}
     \end{subfigure}
     
     \begin{subfigure}{0.24\columnwidth}
        \centering
        \hspace{-0.2cm}\includegraphics[width=1\columnwidth]{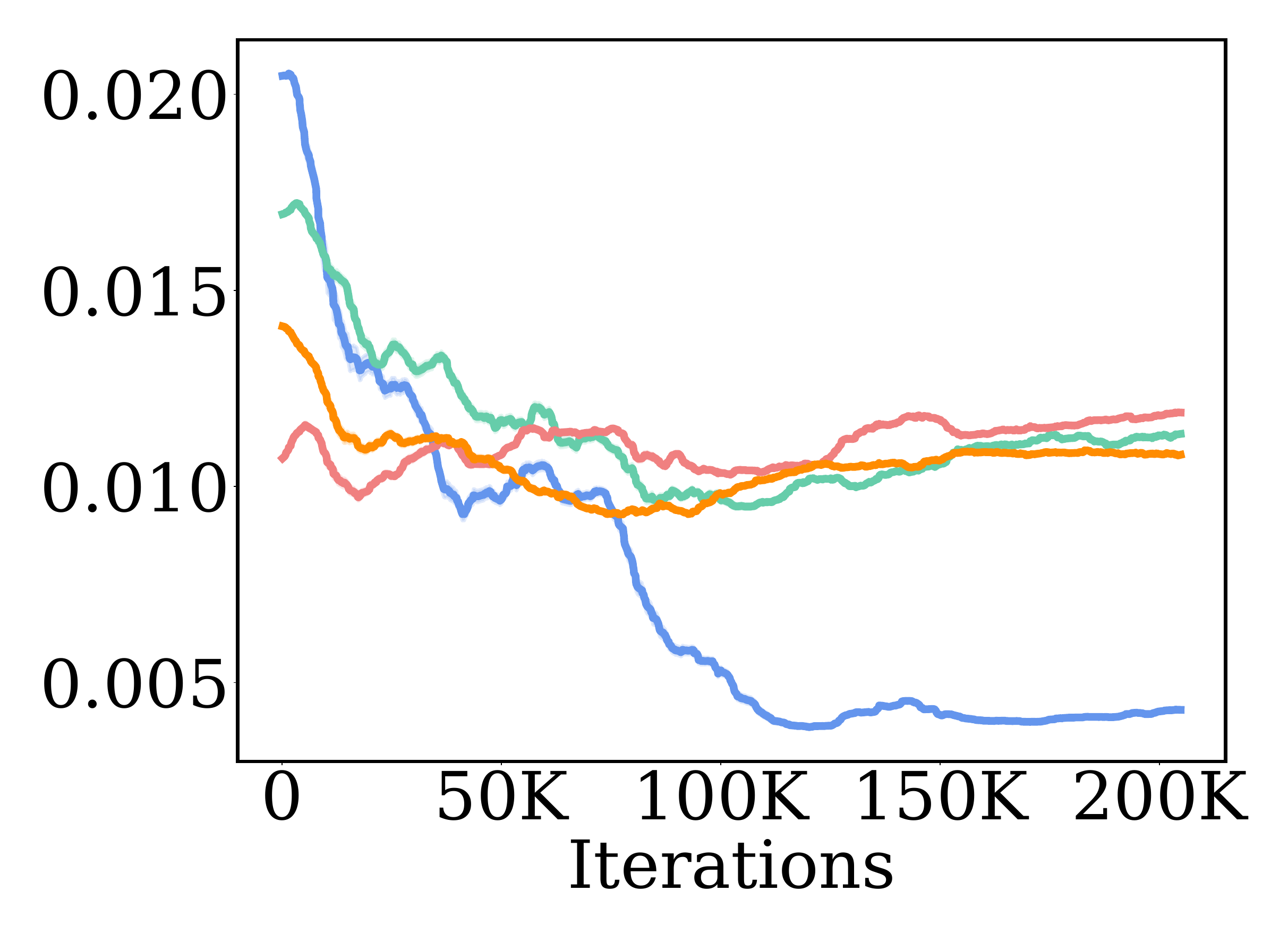}
        \vspace{-0.1cm}
        \caption{}
        \label{fig:4e}
     \end{subfigure}
     \begin{subfigure}{0.24\columnwidth}
        \centering
        \hspace{-0.2cm}\includegraphics[width=1\columnwidth]{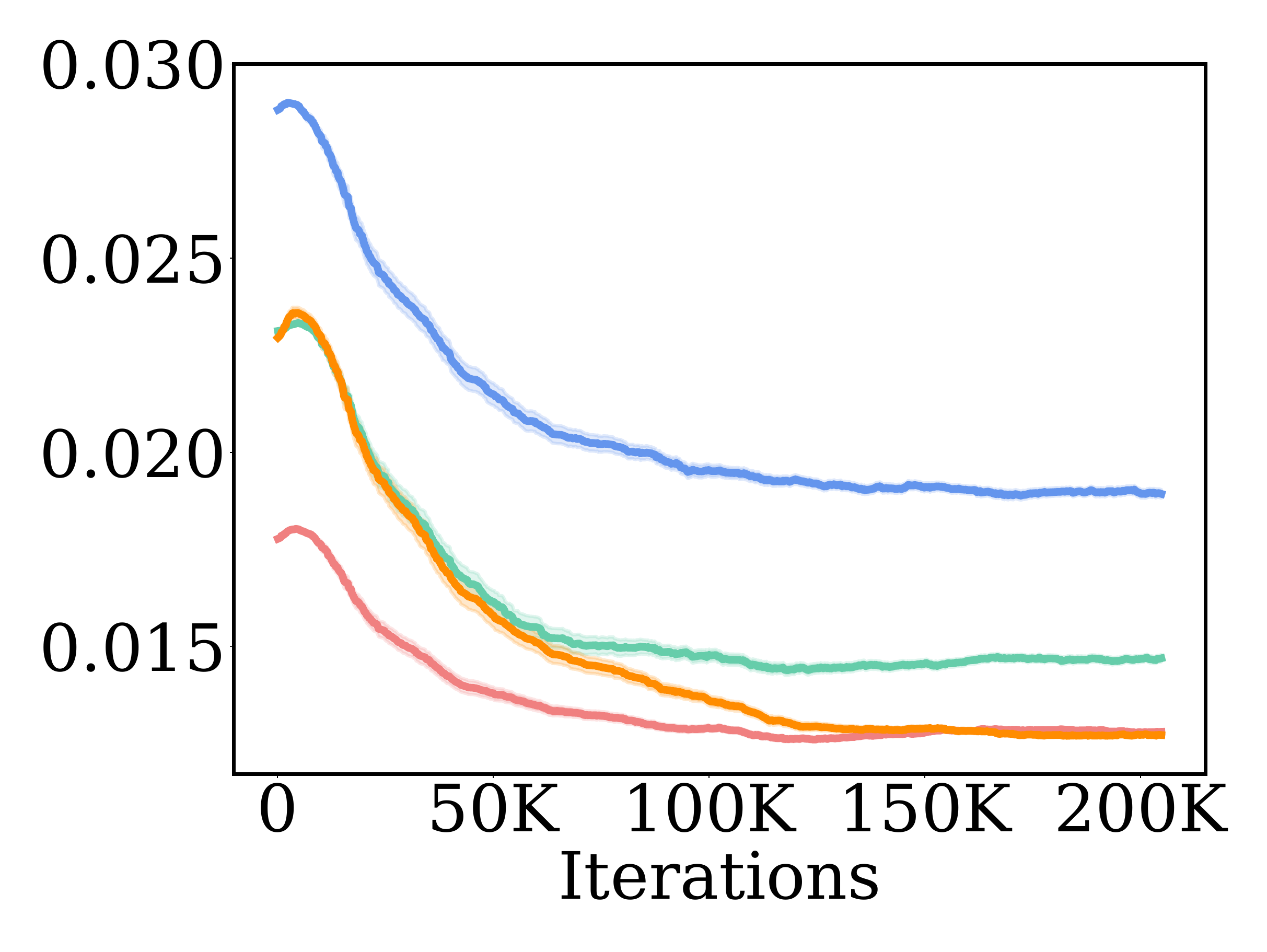}
        \vspace{-0.1cm}
        \caption{}
        \label{fig:4f}
     \end{subfigure}
     \begin{subfigure}{0.24\columnwidth}
        \centering
        \hspace{-0.2cm}\includegraphics[width=1\columnwidth]{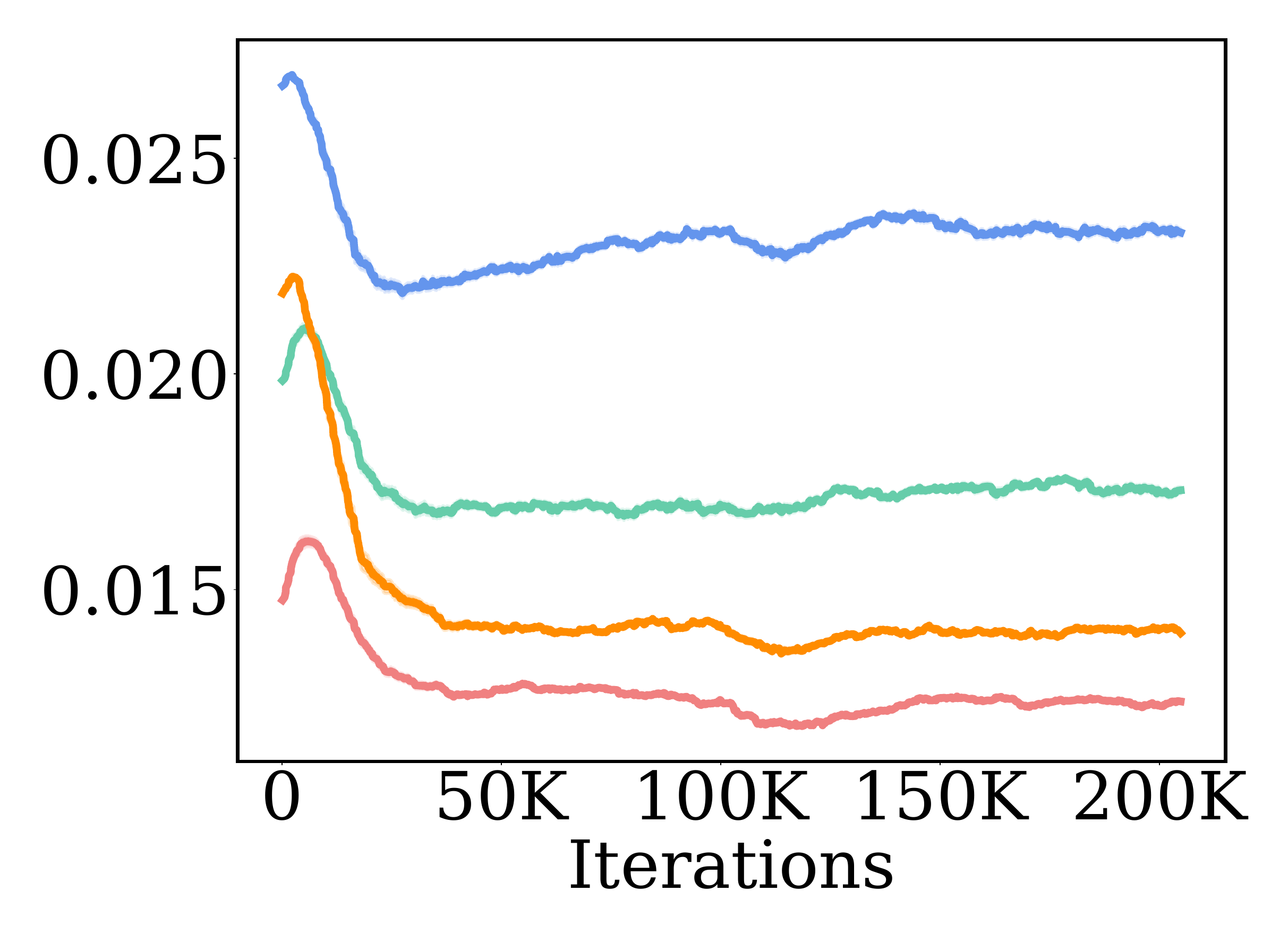}
        \vspace{-0.1cm}
        \caption{}
        \label{fig:4g}
     \end{subfigure}
     \begin{subfigure}{0.23\columnwidth}
        \centering
        \hspace{-0.2cm}\includegraphics[width=1\columnwidth]{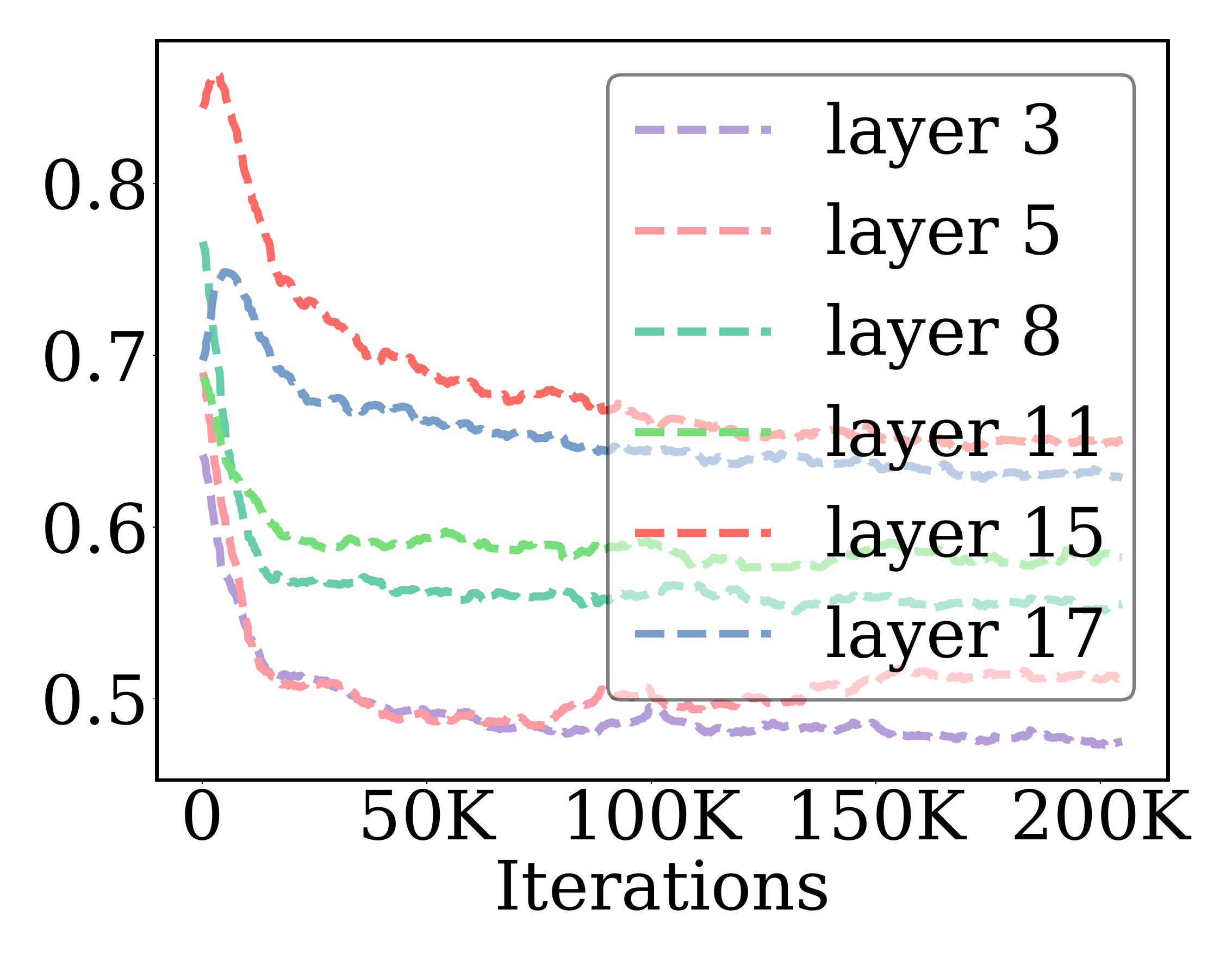}
        \vspace{-0.1cm}
        \caption{}
        \label{fig:4h}
     \end{subfigure}
  \end{center}
  \vspace{-0.7cm}
        \caption{Comparison of ours with the baselines in terms of quantization error for gradients. (a-c) $E(G_{L})$ in 5th, 15th, 17th layers, respectively; (d) Training loss; (e-g) $E(G)$ in 5th, 15th, 17th layers, respectively; (h) Clipping factors. (Best viewed in color.)}
	\vspace{-2mm} 
  \label{fig:itualgo}
\end{figure*}

\begin{figure}[t]
  \vspace{-1mm}
\begin{center}
\begin{subfigure}[b]{0.35\linewidth}
  \centering
  \includegraphics[width=1\linewidth]{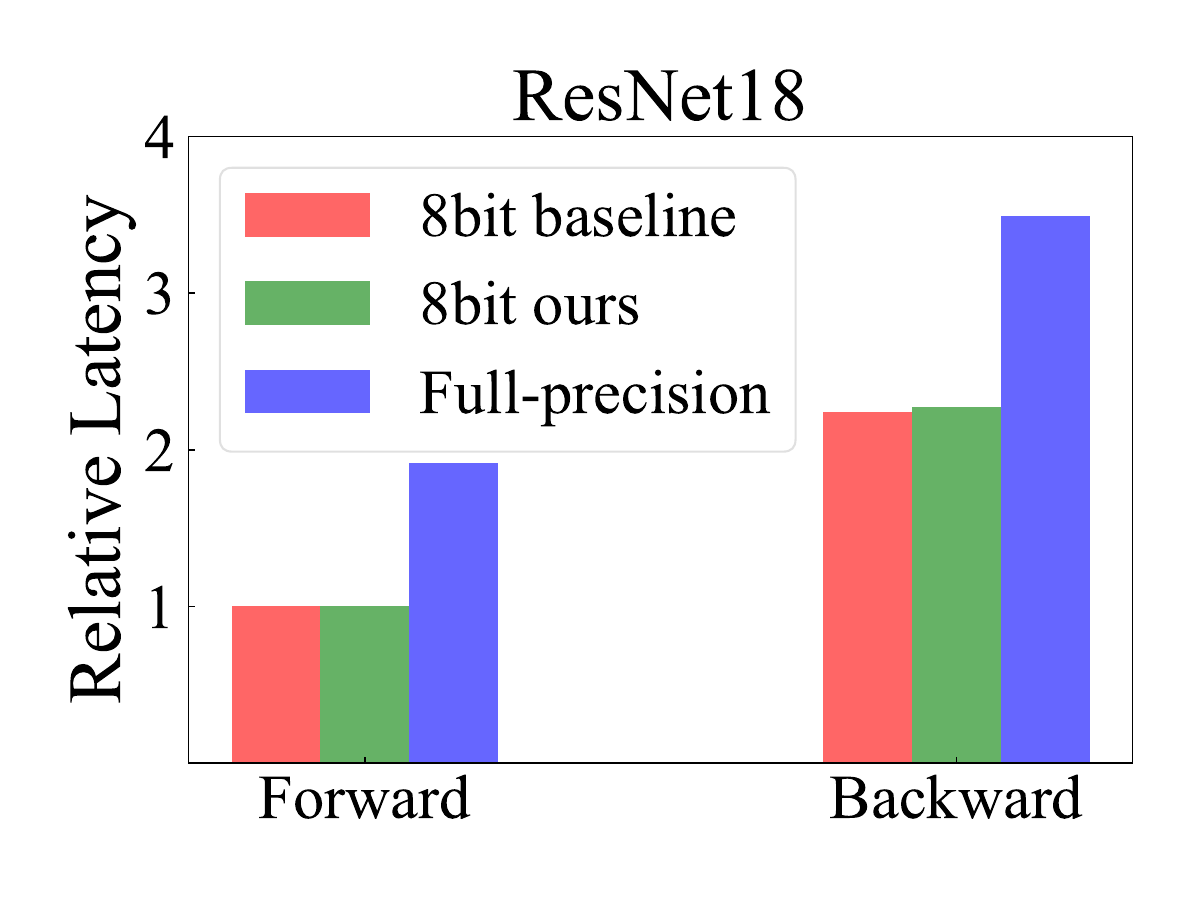}
  \vspace{-7mm}
\end{subfigure}
\begin{subfigure}[b]{0.35\linewidth}
  \centering
  \includegraphics[width=1\linewidth]{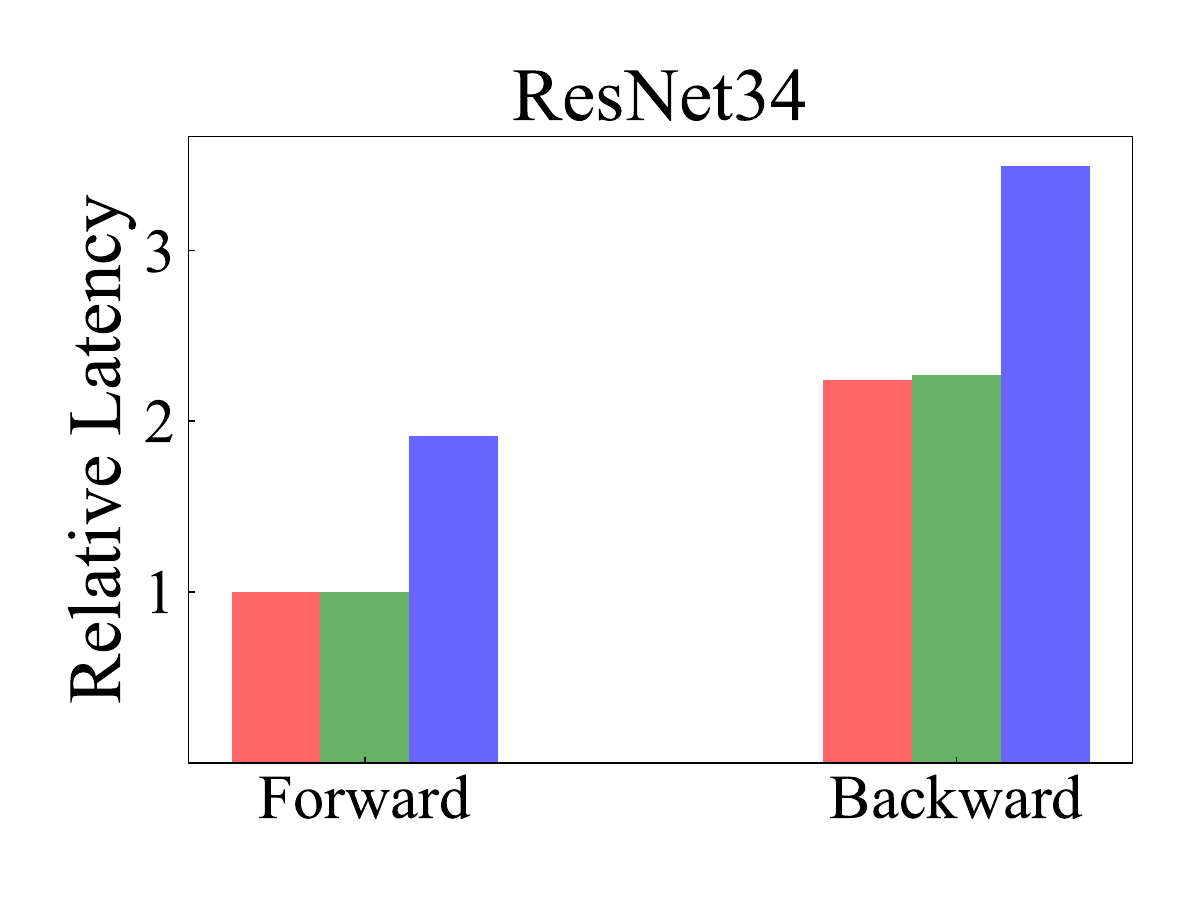}
  \vspace{-7mm}
\end{subfigure}
\vspace{-3mm}
\captionsetup{font={small}}
\caption{Comparisons of latencies for forward and backward passes using TITAN RTX on CIFAR-100~\cite{krizhevsky2009learning}. We normalize the forward latency of baseline to 1.}
\label{fig:train_time}
\end{center}
\vspace{-5mm}
\end{figure}

\paragraph{Analysis on updating intervals.}
We compare in Fig.~\ref{fig:itualgo} our method and baselines with different clipping factors ($\gamma=0.6,0.8,1.0$), in terms of the quantization error for gradients. We train ResNet-20 with the 4/4/4-bit setting on CIFAR-100~\cite{krizhevsky2009learning}. We can see that our method brings a small quantization error for large gradients compared to other baselines, regardless of layers and iterations~(Figs.~\ref{fig:4a},~\ref{fig:4b},~\ref{fig:4c}). This suggests that adjusting the clipping factor according to the condition in Eq.~\eqref{eq:optimal} is effective to maintaining a small error for large gradients. We also compare the quantization error of entire gradients for ours and the baselines~(Figs.~\ref{fig:4e},~\ref{fig:4f},~\ref{fig:4g}). We can observe that performance and the quantization error for entire gradients are less correlated to each other. For example, ours brings a large error for entire gradients compared to the baseline ($\gamma=0.6$) in the 15th and the 17th layers. It also shows a larger error than other baseline ($\gamma=1.0$) in the 5th layer. Nevertheless, our method outperforms the baselines significantly (Table~\ref{tab:small_data}, Fig.~\ref{fig:4d}). This strengthens our motivation that lowering the error for large gradients, rather than entire gradients, plays a crucial role in enhancing the performance of the FXP training. We can see from Fig.~\ref{fig:4h} that the clipping factors vary depending on layers and training iterations, since our update algorithm adjusts them according to the distribution of gradients. For example, if most gradients are concentrated near zero and large gradients are distributed broadly around the tail of the distribution, a small clipping factor is preferred. On the other hand, if a number of large gradients exist in the tail densely, a larger clipping factor would be used. We can also observe that the clipping factors are relatively small in later iterations. A reason is that gradients become sparse and they are around a zero value, as the training progresses, as observed in~\cite{zhu2020towards}. Moreover, the gradients in early layers are likely to be sparse compared to the ones in later layers~\cite{zhu2020towards}, and clipping factors for the early layers thus tend to be small values.                                                                                                                                                                                                                                                                                                  
\vspace{-3mm}
\paragraph{Runtime analysis.}
We compare in Fig.~\ref{fig:train_time} the relative latencies for forward and backward passes. Specifically, we convert the data formats of weights, activations, and gradients to 8-bit and simulate the low-bit operations. We can observe that ours and the baseline, which does not use an interval update algorithm, accelerate the training process in both forward and backward passes compared to the FP models. We can also see that ours and baseline show almost the same latency, demonstrating that our interval update algorithm introduces marginal computations compared to overall convolutional operations of the network.

\begin{table*}[t] 
  \setlength{\tabcolsep}{0.3em}
  \small
  \centering
  \bgroup
  \def\arraystretch{1.1}
  \caption{Quantitative comparison of gradient quantization methods on object detection. We report results on the test split of COCO~\cite{lin2014microsoft} in terms of mAP (averaged over IoU thresholds).}
  \vspace{-0.4cm}
  \label{tab:detection_data}
  \begin{adjustbox}{width=0.8\columnwidth,center}
  \begin{tabular}{C{1.5cm}C{1.9cm}lC{0.9cm}C{2.1cm}C{2.1cm}}
  \hline
  \multicolumn{1}{c}{\multirow{2}{*}{Model}} & \multicolumn{1}{c}{\multirow{2}{*}{Backbone}} &  \multicolumn{1}{c}{\multirow{2}{*}{Method}}& \multicolumn{3}{c}{Bit-width (W/A/G) } \\

    & & & FP & 8/8/8 & 6/6/6\\  
  \hline
  \multicolumn{1}{c}{\multirow{3}{*}{\shortstack[c]{Faster\\R-CNN}}} & \multicolumn{1}{c}{\multirow{3}{*}{\shortstack[c]{ResNet-50}}} & Baseline ($\gamma$=1.0)       		& 38.2 & 36.6 ($-$1.6) & 33.8 ($-$4.4) \\
  & & DSGC~\cite{zhu2020towards}	& 36.9 & 35.1 ($-$1.4) & -  \\
  & & \cellcolor[HTML]{DAE8FC}Ours 	& \cellcolor[HTML]{DAE8FC}38.2 & \cellcolor[HTML]{DAE8FC}\bf{37.0 ($-$1.2)} & \cellcolor[HTML]{DAE8FC}\bf{34.2 ($-$4.0)}  \\
  \hline
  \end{tabular}
  \end{adjustbox}
  \egroup
  \vspace{-0.3cm}
\end{table*}

\vspace{-0.2cm}
\subsubsection{Object detection.} We compare our approach with DSGC~\cite{zhu2020towards} and the baseline ($\gamma=1.0$) on COCO~\cite{lin2014microsoft} that provides over 330,000 images of 80 object categories. We apply ours and the baseline ($\gamma=1.0$) to the Faster R-CNN architecture~\cite{ren2015faster} with 8/8/8 and 6/6/6-bit settings, and then report the performance in terms of mAP in Table~\ref{tab:detection_data}. From the table, we can observe that our method shows better results compared to other methods, confirming the effectiveness of our approach once more. Although the full-precision model for DSGC shows a lower mAP than ours, the performance drop of our method w.r.t the full-precision model is lower, compared to DSGC. This verifies that lowering the quantization error for large gradients is more effective to alleviate performance degradation of low-bit FXP training for object detection, compared to that of entire gradients. We provide qualitative comparisons in the supplementary material.
\begin{table*}[t] 
  \small
  \centering
  \bgroup
  \def\arraystretch{1.1}
  \caption{Quantitative comparison of gradient quantization methods on image super-resolution. We report the average PSNR for different scale factors (2x, 3x, and 4x) on Set5~\cite{bevilacqua2012low}.}
  \vspace{-0.4cm}
  \label{tab:sr_data}
  \begin{adjustbox}{width=0.55\columnwidth,center}
  \begin{tabular}{C{1.9cm}C{0.9cm}lC{1.0cm}C{2.4cm}C{2.4cm}C{2.4cm}}
  \hline
  Architecture & \multicolumn{1}{c}{\multirow{2}{*}{Scale}} & \multicolumn{1}{c}{\multirow{2}{*}{Method}}& \multicolumn{2}{c}{Bit-width (W/A/G) } \\

   (Dataset) & & & FP & 4/4/4\\  
  \hline
  \multicolumn{1}{c}{\multirow{7}{*}{\shortstack[c]{EDSR \\(Set5)}}} & \multicolumn{1}{c}{\multirow{2}{*}{\shortstack[c]{x2}}} & Baseline ($\gamma$=1.0)       		& 38.10 & 37.50 ($-$0.60) \\
  & & \cellcolor[HTML]{DAE8FC}Ours									& \cellcolor[HTML]{DAE8FC}38.10 & \cellcolor[HTML]{DAE8FC}\bf{37.61} ($-$0.49) \\ 
  \cline{2-7}
  & \multicolumn{1}{c}{\multirow{2}{*}{\shortstack[c]{x3}}} & Baseline ($\gamma$=1.0)       		& 34.58 & 34.18 ($-$0.40) \\
  & & \cellcolor[HTML]{DAE8FC}Ours									& \cellcolor[HTML]{DAE8FC}34.58 & \cellcolor[HTML]{DAE8FC}\bf{34.26} ($-$0.32) \\ 
  \cline{2-7}  
  & \multicolumn{1}{c}{\multirow{2}{*}{\shortstack[c]{x4}}} & Baseline ($\gamma$=1.0)       		& 32.31 & 31.82 ($-$0.49) \\
  & & \cellcolor[HTML]{DAE8FC}Ours									& \cellcolor[HTML]{DAE8FC}32.31 & \cellcolor[HTML]{DAE8FC}\bf{31.93} ($-$0.38) \\ 

\hline

  \end{tabular}
  \end{adjustbox}
  \egroup
  \vspace{-4mm}
\end{table*}
\vspace{-0.3cm}
\subsubsection{Image super-resolution.} We apply our method to quantize gradients for EDSR~\cite{lim2017enhanced} on image super-resolution, and demonstrate the generalization ability.  To our knowledge, we are the first to quantize gradients for image super-resolution, making it difficult to compare the performance of ours and existing gradient quantization methods~\cite{yang2020training,zhu2020towards,liu2022IQB} on image super-resolution. We provide a quantitative comparison in Table~\ref{tab:sr_data}, where we report the average PSNR for upsampled images with factors of 2, 3, and 4, on Set5~\cite{bevilacqua2012low}. From this table, we can see that our method provides better results than the baseline regardless of bit-widths, demonstrating that our method is particularly effective in the low-bit gradient quantization. Note that EDSR is trained with the Adam optimizer, in contrast to the networks for image classification and object detection using SGD, suggesting that our method is robust to the type of optimizers. More results on various datasets and qualitative comparisons can be found in the supplementary material.

\vspace{-0.3cm}
\section{Conclusion}
\vspace{-0.3cm}
We have shown an influence of the quantization error for gradients on a low-bit FXP training through experimental analysis, and found that minimizing the quantization error for large gradients contributes to boosting the performance significantly. Based on this, we have introduced the simple yet effective interval update algorithm adjusting the quantization interval adaptively to keep the small quantization error for large gradients. We have presented that our update algorithm achieves the state of the art on the low-bit training for various network architectures and bit-widths. We believe that our approach provides a significant advancement in low-bit FXP training.

\vspace{-0.4cm}
\subsection*{Acknowledgement.}
\vspace{-0.3cm}
This research was supported by Samsung Research Funding \& Incubation Center of Samsung Electronics under Project Number SRFC-IT2102-06.


%
%
\bibliographystyle{splncs04}
\bibliography{main}

\clearpage


\section*{Supplementary material}

# Toward INT4 Fixed-Point Training via Exploring Quantization Error for Gradients Supplement

Dohyung Kim[1], Junghyup Lee[1], Jeimin Jeon[1,2],
Jaehyeon Moon[1,2], and Bumsub Ham[1,⋆]

[1]Yonsei University, [2]Articron Inc.
http://cvlab.yonsei.ac.kr/projects/LBT

## S1  Derivation of Eq. (8)

We show in Fig. A an example of a gradient distribution. To compute ULG, we define upper bounds of the quantization error for clip-in and clip-out gradients: 1) The clip-in gradients are defined as large gradients inside the quantization interval, shown in the blue arrow in Fig. A. We represent the quantization step size (the black arrow in Fig. A) in a $b$-bit setting as $2\gamma g_{max}/(2^b - 2)$. As the gradient moves away from its quantized value, the quantization error increases. Subsequently, the quantization error is maximized at the transition point (the red dash line in Fig. A), where the error quantifies the half of quantization step size. We set the upper bound of quantization error for the clip-in gradient $U_{in}$ to $\gamma g_{max}/(2^b - 2)$. 2) For the clip-out gradient, the quantized value is assigned to each side of the quantization interval (i.e., $\pm\gamma g_{max}$). The quantization error is maximized when the full-precision gradient is located at the minimum or maximum value of gradients, where the error is estimated as $(1 - \gamma)g_{max}$. Using the upper bounds of the error for clip-in and clip-out gradients, we define $U(G_L)$ as follows:

$$
\begin{aligned}
U(G_L) &= \frac{U_{\text{in}}N(G_{\text{in}}, \gamma) + U_{\text{out}}N(G_{\text{out}}, \gamma)}{N(G_L)g_{max}} \\
&= \frac{\frac{\gamma}{2^b-2}g_{max}N(G_{\text{in}}, \gamma) + (1 - \gamma)g_{max}N(G_{\text{out}}, \gamma)}{N(G_L)g_{max}}.
\end{aligned}
\tag{i}
$$

Using $N(G)$, Eq. (i) can be formulated as:

$$
U(G_L) = \frac{\frac{\gamma}{2^b-2}N(G_{\text{in}}, \gamma) + (1 - \gamma)N(G_{\text{out}}, \gamma)}{N(G)} \frac{N(G)}{N(G_L)}.
\tag{ii}
$$

We denote by $\alpha$ the ratio between numbers of large gradients and entire gradients, i.e., $N(G_L)/N(G)$ in Sec. 3.3. Plugging $\alpha$ in Eq. (ii), we obtain the following

---

⋆ Corresponding author

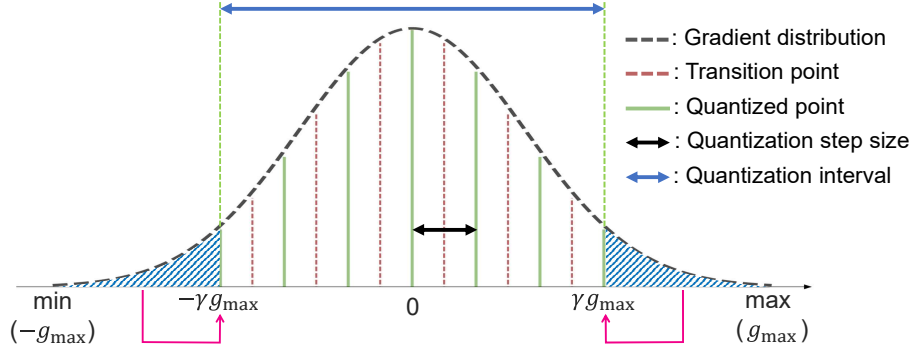

**Fig. A:** Example of a gradient distribution in a 3-bit setting. The gradients located inside the quantization interval (the blue arrow) are mapped to the nearest quantized point (the green lines). The other gradients, located outside the interval, are mapped to end of the interval (the magenta arrows). (Best viewed in color.)

results:

$$U(G_L) = \frac{\frac{\gamma}{2^b-2}N(G_{\text{in}}, \gamma) + (1-\gamma)N(G_{\text{out}}, \gamma)}{N(G)}\frac{1}{\alpha} \tag{iii}$$

$$= \left(\frac{\gamma}{2^b-2}R(G_{\text{in}}, \gamma) + (1-\gamma)R(G_{\text{out}}, \gamma)\right)\frac{1}{\alpha}.$$

## S2 Derivation of Eq. (11)

We take the derivative of Eq. (10) w.r.t the clipping factor $\gamma$ as follows:

$$\frac{dR(G_{\text{in}}, \gamma)}{d\gamma} + \frac{dR(G_{\text{out}}, \gamma)}{d\gamma} = \frac{d\alpha}{d\gamma} = 0. \tag{iv}$$

Note that $\alpha$ is a hyperparameter in our framework, regarded as the constant value. Therefore, the derivative of $\alpha$ w.r.t. the clipping factor $\gamma$ becomes zero, *i.e.*, $d\alpha/d\gamma = 0$. Equation (iv) can then be reformulated as

$$\frac{dR(G_{\text{in}}, \gamma)}{d\gamma} = -\frac{dR(G_{\text{out}}, \gamma)}{d\gamma}. \tag{v}$$

We compute the derivative of ULG in Eq. (iii) w.r.t. the clipping factor as follows:

$$\frac{dU(G_L)}{d\gamma} = \left(\frac{1}{2^b-2}R(G_{\text{in}}, \gamma) + \frac{\gamma}{2^b-2}\frac{dR(G_{\text{in}}, \gamma)}{d\gamma} - R(G_{\text{out}}, \gamma) + (1-\gamma)\frac{dR(G_{\text{out}}, \gamma)}{d\gamma}\right)\frac{1}{\alpha}. \tag{vi}$$

Plugging Eq. (v) into Eq. (vi), we can obtain the following results:

$$\frac{dU(G_L)}{d\gamma} = \left(\frac{1}{2^b-2}R(G_{\text{in}}, \gamma) - R(G_{\text{out}}, \gamma) + \left(1-\gamma-\frac{\gamma}{2^b-2}\right)\frac{dR(G_{\text{out}}, \gamma)}{d\gamma}\right)\frac{1}{\alpha}, \tag{vii}$$

which corresponds to Eq. (11).

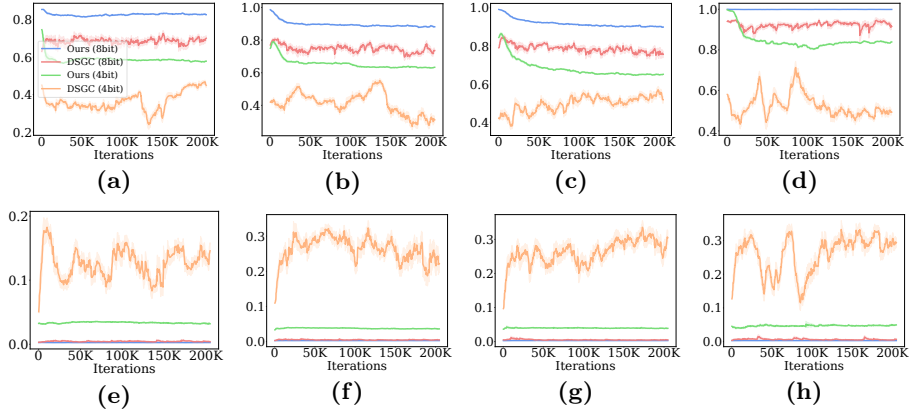

**Fig. B:** Comparison of ours and DSGC in terms of the clipping factor and quantization error for large gradients. (a), (b), (c), (d) Clipping factors $\gamma$ in 7th, 12th, 15th, 18th layers, respectively; (e), (f), (g), (h) $E(G_L)$ in 7th, 12th, 15th, 18th layers, respectively. (Best viewed in color.)

**Table A:** Quantitative comparisons of our method with different values of (a) $\alpha$ and (b) $\beta$ for ResNet-20 on the validation split of CIFAR-100 [4].

| $\alpha$ | 2e-3 | 1.5e-3 | 1e-3 | 0.5e-3 | 1e-4 |
|---|---|---|---|---|---|
| Top-1 Acc. | 63.2 | 63.24 | **63.4** | 61.33 | 61.03 |

**(a)**

| $\beta$ | 1e-2 | 1e-3 | 1e-4 | 1e-5 |
|---|---|---|---|---|
| Top-1 Acc. | 63.02 | **63.4** | 63.11 | 63.33 |

**(b)**

**Table C:** Quantitative comparisons of our method and baseline ($\gamma = 1.0$) for ViT-L [2] on the test split of CIFAR-10 [4]. W/A/G: Bit-precision of weights/activations/gradients; FP: Results obtained by full-precision models.

| Method | FP | 6/6/6 | 5/5/5 | 4/4/4 |
|---|---|---|---|---|
| Baseline ($\gamma = 1.0$) | 97.2 | 94.80 (-2.40) | 89.36 (-7.84) | 83.40 (-13.8) |
| Ours | 97.2 | **97.14 (-0.06)** | **97.12 (-0.08)** | **96.12 (-1.08)** |

## S3    Comparison with DSGC

We show in Fig. B a comparison of our method and DSGC in terms of the clipping factor and quantization error for large gradients in 8-bit or 4-bit settings. From this, we observe three things: (1) Both DSGC and our method show lower quantization errors for large gradients in the 8-bit setting, and their performances in the setting are comparable to the full-precision counterparts (Table 1). Note that ours brings a slightly lower error compared to DSGC in the 8-bit setting, providing better quantization performance. (2) For DSGC, the quantization error for large gradients increases drastically as the bit-width is reduced from 8-bit to 4-bit settings. This is consistent with the reason why the accuracy of DSGC decreases severely when the bit-width is reduced, as described in Sec. 4.2. (3) We can observe that our method reduces the clipping factors as the bit-width

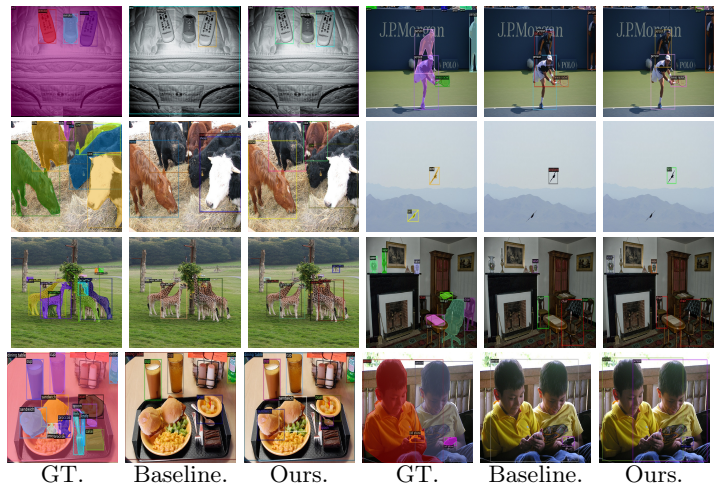

**Fig. C:** Visual comparison of detection results on the COCO dataset [6]. We quantize the Faster R-CNN model [8] using the baseline and our method, where weight, activation, and gradient are quantized to 6-bit fixed-point values. (Best viewed in color.)

decreases. The clipping factors in the 4-bit setting tend to be lower than those in the 8-bit setting. According to the decrement in the bit-width, optimal values for the clip-in and clip-out ratios in Eq. (14) decreases and increases, respectively. The clipping factor thus decreases to satisfy the conditions for the clip-in and clip-out ratios.

## S4     Hyperparameter search

We show in Table A the quantitative results on CIFAR-100 [4] using the ResNet-20 architecture with various hyperparameters $\alpha$ and $\beta$. We can observe that setting both $\alpha$ and $\beta$ to 1e-3 provides the best performance. Although searching for optimal $\alpha$ and $\beta$ for each architecture can provide better results, searching costs are not marginal. In this regard, fixing $\alpha$ and $\beta$ via a grid search provides a good compromise between efficiency and efficacy.

## S5     Transformer architecture

We show in Table C the comparisons between our approach and the baseline ($\gamma = 1.0$) using ViT-L [2] on CIFAR-10 [4]. We can see that our method outperforms the baseline across various bit-widths consistently. The baseline shows severe performance degradation in low-bit settings (*e.g.*, 4/4/4 and 5/5/5 bits), but ours maintains the performance closer to that of the FP counterparts. This demonstrates our method can also be applicable and effective in the transformer architecture.

**Table D:** Quantitative comparison of gradient quantization methods on image super-resolution. We report the average PSNR for different scale factors (2x, 3x, and 4x) on Set14 [9], BSD100 [7], and Urban100 [3]. W/A/G: Bit-precision of weights/activations/gradients; FP: Results obtained by full-precision models.

| Architecture (Dataset) | Scale | Method | Bit-width (W/A/G) | |
|---|---|---|---|---|
| | | | FP | 4/4/4 |
| EDSR (Set14) | x2 | Baseline ($\gamma$=1.0) | 33.80 | 33.13 (−0.67) |
| | | Ours | 33.80 | **33.19 (−0.61)** |
| | x3 | Baseline ($\gamma$=1.0) | 30.43 | 30.15 (−0.28) |
| | | Ours | 30.43 | **30.18 (−0.25)** |
| | x4 | Baseline ($\gamma$=1.0) | 28.73 | 28.35 (−0.38) |
| | | Ours | 28.73 | **28.39 (−0.34)** |
| EDSR (BSD100) | x2 | Baseline ($\gamma$=1.0) | 32.23 | 31.92 (−0.31) |
| | | Ours | 32.23 | **31.98 (−0.25)** |
| | x3 | Baseline ($\gamma$=1.0) | 29.17 | 28.96 (−0.21) |
| | | Ours | 29.17 | **28.99 (−0.18)** |
| | x4 | Baseline ($\gamma$=1.0) | 27.65 | 27.43 (−0.22) |
| | | Ours | 27.65 | **27.44 (−0.21)** |
| EDSR (Urban100) | x2 | Baseline ($\gamma$=1.0) | 32.45 | 31.26 (−1.19) |
| | | Ours | 32.45 | **31.42 (−1.03)** |
| | x3 | Baseline ($\gamma$=1.0) | 28.51 | 27.69 (−0.82) |
| | | Ours | 28.51 | **27.79 (−0.72)** |
| | x4 | Baseline ($\gamma$=1.0) | 26.34 | 25.67 (−0.67) |
| | | Ours | 26.34 | **25.69 (−0.65)** |

## S6   Object detection

We present a visual comparison of detection results in Fig. C. We can see that our method performs better than the baseline in terms of localization accuracy. For example, it accurately localizes the objects even in the presence of background clutter and occluded parts (the first row), and labels the objects correctly compared to the baseline (the second row). Our method is also robust to localizing small and/or overlapping objects, compared to the baseline, as shown in the third and fourth rows, respectively.

## S7   Image super-resolution

We provide a quantitative comparison in Table D, where we report the average PSNR for upsampled images with factors of 2, 3, and 4, on Set14 [9], BSD100 [7], and Urban100 [3]. We can observe that ours achieves superior results over the baseline regardless of datasets and bit-widths, verifying the effectiveness of our method once more. We also show a visual comparison of ours and the baseline in Fig. D. We can clearly see that our method provides more sharp results, compared to the baseline, without the aliasing artifact. For example, it accurately reconstructs fine-grained details such as straight lines (*e.g.*, the wooden deck in the first row), textures (*e.g.*, the hijab in the second row), and small-scale structures (*e.g.*, the ornament in the third row).

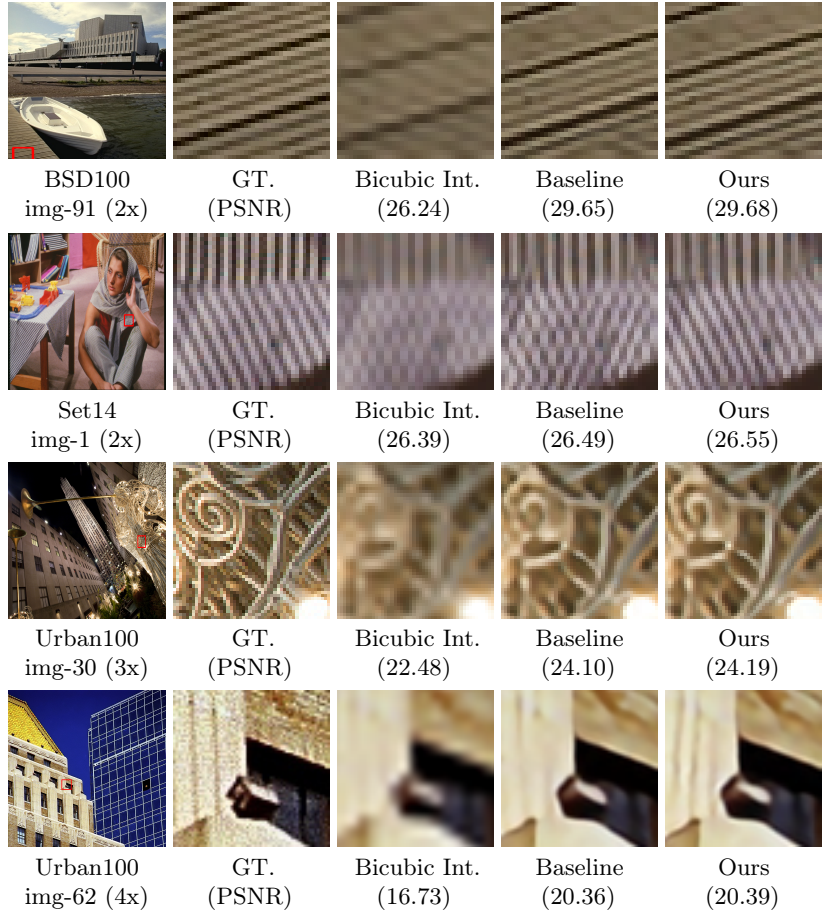

**Fig. D:** Visual comparison of reconstructed HR images (2x, 3x, and 4x) on Set14 [9], Urban100 [3], and BSD100 [7]. We report the average PSNR in the parentheses. We train EDSR [5] on the DIV2k dataset [1] using the baseline and our method, with the 4/4/4-bit setting. (Best viewed in color.)

## S8    Algorithm table.

We summarize the process of updating the clipping factor in the Algorithm. 1.

---

**Algorithm 1** Clipping factor update.

---

1: **Hyperparameter**: $b$: a bit-width for gradient quantization; $\alpha$: a large gradients ratio; $\beta$: a scaling parameter.

2: **Input**: $\gamma_{i-1}$: a clipping factor at the $(i-1)$-th iteration; $G$: a set of entire gradients; $G_{\text{out}}$: a set of clip-out gradients.

3: **Output**: $\gamma_i$: a clipping factor at the $i$-th iteration.

4: **Clipping factor update**:

5: Compute the clip-out ratio $R(G_{out}, \gamma_{i-1})$: $R(G_{\text{out}}, \gamma_{i-1}) = \frac{N(G_{\text{out}}, \gamma_{i-1})}{N(G)}$, where $N(G)$ and $N(G_{\text{out}}, \gamma_{i-1})$ count the number of elements in the set of $G$ and $G_{\text{out}}$, respectively.

6: Compute the direction of updating the clipping factor: $\text{sign}(T(G_{\text{out}}, \gamma_{i-1})) = \text{sign}(R(G_{\text{out}}, \gamma_{i-1}) - \frac{1}{2^b-1}\alpha)$, where $\text{sign}(\cdot)$ is a signum function.

7: Update the clipping factor at the $(i-1)$-th iteration: $\gamma_i = \gamma_{i-1} + \beta \text{sign}(T(G_{\text{out}}, \gamma_{i-1}))$

---



\end{document}